\documentclass[11pt]{article}

\usepackage[final]{acl}
\usepackage{times}
\usepackage{latexsym}
\usepackage[T1]{fontenc}
\usepackage[utf8]{inputenc}
\usepackage{microtype}
\usepackage{inconsolata}
\usepackage{graphicx}

\usepackage{amsmath}
\usepackage{amssymb}
\usepackage{booktabs}
\usepackage{multirow}
\usepackage{colortbl}
\usepackage{algorithm}
\usepackage{algpseudocode}
\usepackage{subcaption}
\usepackage{adjustbox}

\usepackage{pifont}
\newcommand{\cmark}{\textcolor{GoodGreen}{\ding{51}}}
\newcommand{\xmark}{\textcolor{BadRed}{\ding{55}}}
\usepackage[most]{tcolorbox}
\usepackage{listings}
\usepackage{fancyvrb}
\usepackage{fvextra}
\usepackage{pifont}
\usepackage{xcolor}
\usepackage{booktabs}
\usepackage{makecell}
\definecolor{promptgreen}{HTML}{12A312}   
\definecolor{promptbg}{HTML}{ebfff2}   
\definecolor{prompttext}{HTML}{6e6e6e}    
\definecolor{GoodGreen}{HTML}{2EAD63}
\definecolor{BadRed}{HTML}{E74C3C}
\definecolor{LightBlue}{HTML}{EAF3FF}

\lstdefinestyle{obs}{basicstyle=\ttfamily\normalsize,breaklines=true,
  columns=fullflexible,keepspaces=true,xleftmargin=8pt,
  commentstyle=\itshape\color{gray!70}}
\newtcolorbox{issuebox}{colback=yellow!8,colframe=yellow!55!black,boxrule=.8pt,
  arc=2pt,left=10pt,right=10pt,top=6pt,bottom=6pt,
  fonttitle=\bfseries\normalsize,title={Issue: sphinx-doc/sphinx\#10449}}
\newtcolorbox{actbox}[1]{colback=blue!5,colframe=blue!55!black,boxrule=.8pt,
  arc=2pt,left=10pt,right=10pt,top=5pt,bottom=5pt,
  fonttitle=\bfseries\normalsize,title={Action #1}}
\newtcolorbox{obsbox}[1]{colback=gray!8,colframe=gray!55,boxrule=.8pt,
  arc=2pt,left=10pt,right=10pt,top=5pt,bottom=5pt,
  fonttitle=\bfseries\normalsize,title={Observation #1}}
\newtcolorbox{finishbox}{colback=green!5,colframe=green!55!black,boxrule=.8pt,
  arc=2pt,left=10pt,right=10pt,top=5pt,bottom=5pt,
  fonttitle=\bfseries\normalsize,title={Finish}}
\newtcolorbox{finishboxfail}{colback=red!5,colframe=red!55!black,boxrule=.8pt,
  arc=2pt,left=10pt,right=10pt,top=5pt,bottom=5pt,
  fonttitle=\bfseries\normalsize,title={Finish}}

\lstdefinestyle{promptstyle}{
    basicstyle=\ttfamily\fontsize{9.8pt}{12.2pt}\selectfont\color{prompttext},
    breaklines=true,
    columns=fullflexible,
    keepspaces=true,
    showstringspaces=false,
    frame=none,
    aboveskip=0pt,
    belowskip=0pt
}

\lstdefinestyle{obs}{basicstyle=\ttfamily\small,breaklines=true,
  columns=fullflexible,keepspaces=true,xleftmargin=0pt,
  commentstyle=\itshape\color{gray!70}}

\newtcblisting{promptbox}[1]{
    width=\textwidth,
    colback=promptbg,
    colframe=promptgreen,
    colbacktitle=promptgreen,
    coltitle=white,
    coltext=prompttext,
    title={#1},
    fonttitle=\normalfont\fontsize{11.2pt}{12.5pt}\selectfont,
    listing only,
    listing style=promptstyle,
    arc=1.3mm,
    boxrule=0.7pt,
    left=3mm,
    right=3mm,
    top=1.5mm,
    bottom=1.5mm,
    before skip=2mm,
    after skip=1mm,
    listing options={style=promptstyle}
}

\newcommand{\issueline}{\noindent\rule{\linewidth}{1pt}}
\newcommand{\turnline}{\vspace{2pt}\noindent\rule{\linewidth}{0.4pt}\vspace{2pt}}

\lstdefinestyle{patchpromptstyle}{
    basicstyle=\ttfamily\fontsize{9.8pt}{12.2pt}\selectfont\color{patchprompttext},
    breaklines=true,
    columns=fullflexible,
    keepspaces=true,
    showstringspaces=false,
    frame=none,
    aboveskip=0pt,
    belowskip=0pt
}

\newtcblisting{patchpromptbox}[1]{
    width=\textwidth,
    colback=patchpromptbg,
    colframe=patchpromptmain,
    colbacktitle=patchpromptmain,
    coltitle=white,
    coltext=patchprompttext,
    title={#1},
    fonttitle=\normalfont\fontsize{11.2pt}{12.5pt}\selectfont,
    listing only,
    listing style=patchpromptstyle,
    arc=1.3mm,
    boxrule=0.7pt,
    left=3mm,
    right=3mm,
    top=1.5mm,
    bottom=1.5mm,
    before skip=2mm,
    after skip=1mm,
    listing options={style=patchpromptstyle}
}

\title{\textsc{$\text{A}^2$Agent}: Action-Aware Reinforcement Learning for Repository-Level Code Localization Agents}

\author{
 \textbf{Doyeon Kim}\thanks{Equal contribution},
 \textbf{Suyoung Bae}\footnotemark[1],
 \textbf{Yumin Lee},
 \textbf{Jee-Hyong Lee}\thanks{Corresponding author}
\\
 College of Computing and Informatics \\ Sungkyunkwan University, South Korea
\\
  \{kimdozz0, sybae01, totoandy, john\}@skku.edu
}

\begin{document}
\maketitle

\begin{abstract}
Localizing issue-relevant code regions is a critical step in automated software engineering. However, due to their reliance on sparse trajectory-level signals, existing methods cannot identify which per-turn actions are effective and often discover correct code regions during exploration but fail to commit them. To address these limitations, we propose an action-aware reinforcement learning method that combines a \textit{per-turn reward sequence} rewarding both the discovery and commitment of gold code regions with an \textit{action-level advantage estimation} scheme that isolates each action's credit by grouping turns sharing the same exploration context. Extensive evaluations show that our method improves the average F1 over the state-of-the-art (SOTA) by 1.58\% on SWE-Bench Verified and 8.55\% on SWE-Bench Pro, with our 4B model outperforming baselines up to $8\times$ larger. Our code is available at \url{https://github.com/donian00/A2Agent}.
\end{abstract}

\section{Introduction}
\label{sec:intro}
Recent advances in large language models (LLMs) have shown strong performance across a wide range of code-related tasks~\cite{chen2021evaluating, roziere2023code, guo2024deepseek}. Among them, code localization, which identifies the specific code regions that must be modified to resolve a real software issue, has become a central research focus. Accurate localization is a prerequisite for any downstream issue resolution, and inaccurate localization fundamentally limits the quality of the resulting issue resolution~\cite{jimenez2024swebench}. 

As modern software repositories grow in scale and dependency complexity, a single LLM inference is no longer sufficient to localize all relevant code regions. As a result, agent-based approaches have emerged as the dominant paradigm, where models iteratively invoke tools to navigate the codebase~\cite{chen2025locagent, yu2025orcaloca, jiang2025cosil, zhang2025one}.

\begin{figure}[t]
\centering
\includegraphics[width=\columnwidth]{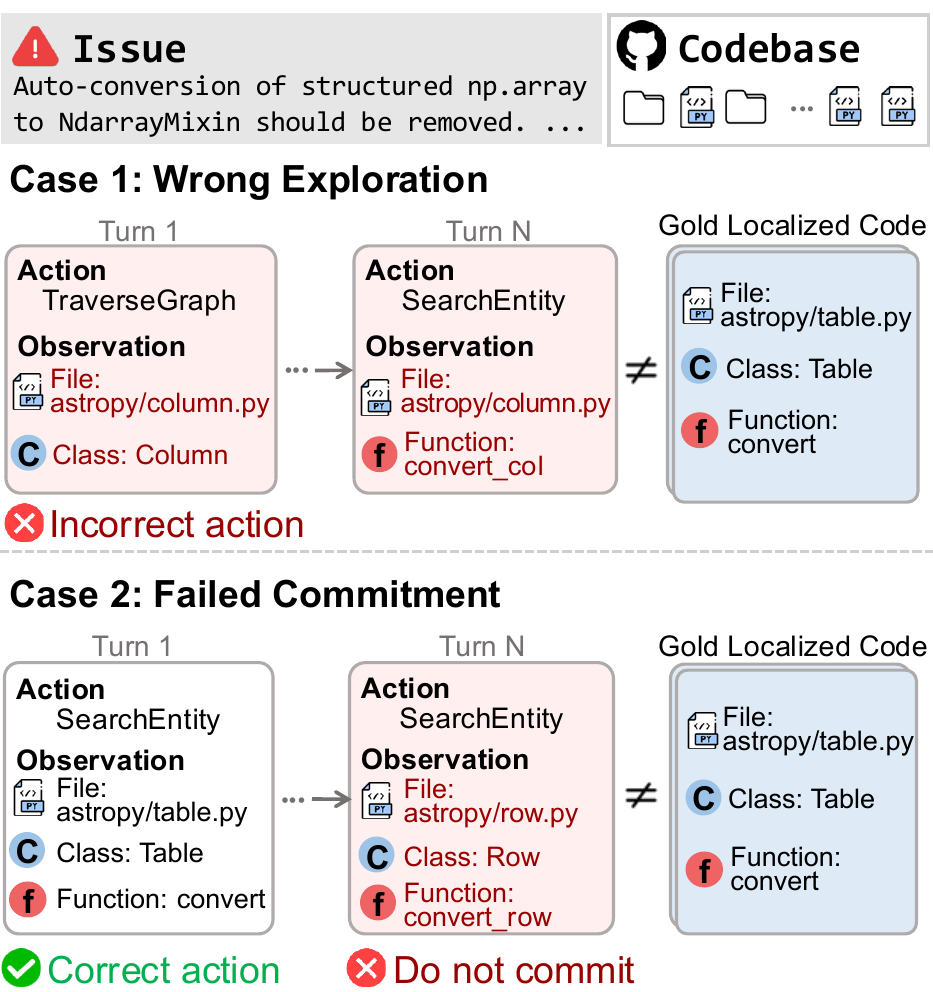}
\caption{Failure cases of previous agentic LLMs for repository-level code localization. \textbf{(Case 1) Wrong exploration}: Agents cannot identify effective actions at each turn, causing unnecessary tool calls that do not contribute to issue resolution. \textbf{(Case 2) Failed commitment}: Even though agents observe the gold region during localization, they fail to recognize and commit it to the final prediction.}
\label{fig:motivation}
\vspace{-0.4cm}
\end{figure}

Early approaches leverage few-shot prompting to elicit tool use and multi-step reasoning from pre-trained LLMs~\citep{yao2022react, wei2022chain}. However, such demonstrations are insufficient for learning complex multi-turn tool use. Therefore, recent work has shifted toward post-training paradigms, training code localization agents via supervised fine-tuning (SFT)~\citep{chen2025locagent, ma2025tool, pan2024training} or reinforcement learning with verifiable rewards (RLVR)~\citep{zhang2025one, sutawika2026codescout, wei2026swe}. However, several limitations remain unresolved by these approaches, as illustrated in Figure~\ref{fig:motivation}.

Existing methods fail to distinguish the contribution and error of each action taken at individual turns, resulting in redundant tool calls that do not contribute to issue resolution. 
In multi-turn code localization, actions at each turn may either contribute to identifying the correct code regions or mislead the search through inappropriate tool calls~\citep{wang2025improving, yu2025orcaloca}. However, existing methods rely on either a single gold trajectory or a sparse outcome-level signal, which cannot separate correct actions from erroneous ones. The agent thus cannot reinforce effective exploration or discount counterproductive behavior.

Moreover, in multi-turn code localization, the agent may observe the correct code regions during exploration but fail to commit them to the final prediction, instead invoking additional tools and outputting incorrect regions (analysis illustrated in Figure~\ref{fig:dc_gap}). Such failures persist because existing trajectory-level sparse rewards treat these explored-but-uncommitted trajectories the same as ones that never reached the correct elements. The discovery–commit gap thus remains unresolved.

To address these challenges, we propose \textbf{\textsc{$\text{A}^2$Agent}}, an \textbf{\textsc{A}}ction-\textbf{\textsc{A}}ware reinforcement learning method for repository-level code localization \textbf{\textsc{Agent}}s. To strengthen the agent's ability to perform effective exploration and reliable commitment, we first design a \textit{per-turn reward sequence} that rewards the discovery and commitment of gold code regions, providing learning signals from partial successes. Second, we propose an \textit{action-level advantage estimation} scheme that groups turns sharing the same exploration context, isolating the credit of each action under the same context. Together, these two components enable action-aware RL that optimizes the agent at the action level.

We evaluate \textsc{$\text{A}^2$Agent} on two widely-used benchmarks, SWE-Bench Verified~\cite{jimenez2024swebench} and SWE-Bench Pro~\cite{deng2025swe}, both comprising real-world GitHub issues with their corresponding repositories.
Our experimental results demonstrate that our \textsc{$\text{A}^2$Agent} outperforms the SOTA baselines by 1.58\% and 8.55\% in average F1, respectively. Notably, with Qwen3-4B as the backbone, \textsc{$\text{A}^2$Agent} surpasses baselines trained on models up to $8\times$ larger, suggesting that fine-grained action-level supervision is a more critical performance factor than model scale in repository-level code localization.

\section{Related Work}

\paragraph{Repository-Level Code Localization.}
Early works in repository-level code localization rank files by similarity to the issue description using BM25~\citep{robertson2009probabilistic} or dense retrieval~\citep{reddy2025swerank, suresh2025cornstack}. However, the semantic gap between natural language issues and code, together with the difficulty of capturing inter-file dependencies~\citep{gupta2025sacl, ouyang2025repograph}, motivates agent-based approaches that iteratively invoke tools to navigate the codebase. Subsequent work enriches these agents through graph modeling, symbol tracing, terminal navigation, and commit history~\citep{chen2025locagent, zhang2025one, sutawika2026codescout, wang2025improving, jiang2025cosil}, while pipeline-based methods such as Agentless~\citep{xia2024agentless} take a complementary route by decomposing localization into sequential LLM prompts. Yet existing methods predominantly focus on expanding the search space and collecting relevant context, leaving two key capabilities unsolved: identifying which per-turn actions meaningfully contribute to issue resolution, and committing observed code elements to the final prediction.

\begin{figure*}[t]
\centering
\includegraphics[width=\textwidth]{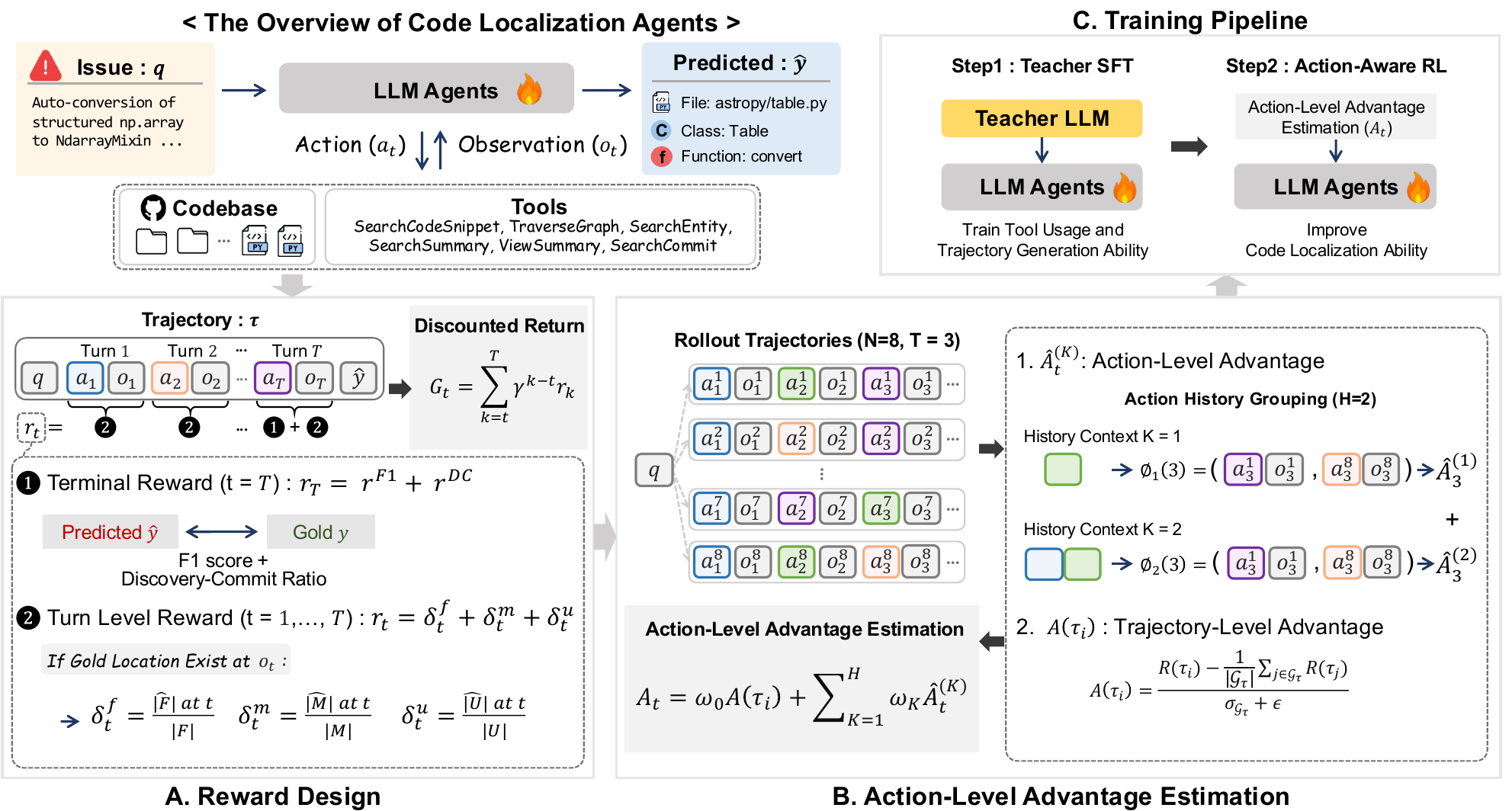}
\caption{Overview of \textsc{$\text{A}^2$Agent}. Given an issue $q$, an LLM agent interacts with the codebase via tool calls to predict the gold target $\hat{y}$. \textit{(A) Reward Design} combines a turn-level reward and a terminal reward to form the discounted return. \textit{(B) Action-Level Advantage Estimation} groups turns sharing the same length-$K$ action history across rollouts to compute a context-matched baseline, producing action-level advantages $\hat{A}^{(K)}_t$ which are then combined with the trajectory-level advantage $A(\tau_i)$. \textit{(C) Training Pipeline} first fine-tunes the agent for tool-use ability, and then optimizes it via Action-Aware RL to improve localization performance.}

\label{fig:method}
\end{figure*}

\paragraph{Agentic Training.}
Early agentic training relied primarily on SFT to imitate teacher trajectories, but it struggles to generalize to unfamiliar codebases at inference time~\cite{deng2025swe, chu2025sft}. Recent work has explored group-based RL methods such as GRPO~\cite{shao2024deepseekmath} and GSPO~\cite{zheng2025group}, which estimate a single trajectory-level advantage from rollout group reward statistics without a separate value function, and have been widely adopted across tasks including code localization~\cite{zhang2025one, sutawika2026codescout, lee2026learning}. However, these methods assign uniform advantages across the entire trajectory based solely on the final prediction, making it difficult to attribute which actions contributed to the final outcome across multiple turns. Another line of work pursues finer-grained credit assignment, estimating advantages at intermediate steps via resampling~\cite{kazemnejad2024vineppo, lee-etal-2026-expo, guo2026segment} or anchoring rollout groups to shared environment states~\cite{feng2026group}. In code exploration, however, the repository remains fixed and multi-turn rollouts are costly to re-execute, leaving no cheap anchor for either approach. We tackle this by leveraging exploration history as a state proxy to compare tool calls within the same search context, enabling action-level advantage estimation at each turn.
\section{Preliminary}
\label{sec:preliminary}
\paragraph{Problem Setup.} We formulate repository-level code localization as a
multi-turn decision-making process. At each turn $t$, the agent observes a state
$s_t$ encoding the issue, repository context, and interaction history, and
produces an action $a_t \sim \pi_\theta(\cdot \mid s_t)$ from a policy $\pi_\theta$.
The agent outputs a localization decision at the final turn $T$, yielding a
trajectory $\tau$. In the conventional setting, a reward is observed only at this final turn, evaluating the final prediction. The discounted return from turn
$t$ is defined as $G_t = \sum_{k=t}^{T} \gamma^{k-t} r_k$, where $\gamma\in(0,1]$.

\paragraph{Group-Relative Advantage Estimation.} Group-based reinforcement learning methods, such as GRPO~\citep{shao2024deepseekmath}, estimate advantages without a separate value function by subtracting the mean return within a group of trajectories $\mathcal{G}_\tau$ sampled for the same issue:
\begin{equation}
    A(\tau_i) = \frac{R(\tau_i) - \frac{1}{|\mathcal{G}_\tau|}\sum_{j \in \mathcal{G}_\tau} R(\tau_j)}{\sigma_{\mathcal{G}_\tau} + \epsilon}.
    \label{eq:grpo}
\end{equation}
Since this value reflects only the relative performance of the entire trajectory, the same advantage is assigned to every turn, making it impossible to reflect the contribution of individual actions. To estimate action-level advantages, one must define the discounted return $G_t$ at turn $t$ and compare turns that took different actions under the same state $\tilde{s}$:

\begin{equation}
    A(\tilde{s}, a) = \frac{G_t - \frac{1}{|\mathcal{G}_{\tilde{s}}|}\sum_{j \in \mathcal{G}_{\tilde{s}}} G_{t_j}}{\sigma_{\mathcal{G}_{\tilde{s}}} + \epsilon}.
    \label{eq:action_advantage}
\end{equation}

\section{Proposed Method}

In this section, we introduce \textbf{\textsc{$\text{A}^2$Agent}}, an \textbf{\textsc{A}}ction-\textbf{\textsc{A}}ware reinforcement learning method for repository-level code localization \textbf{\textsc{Agent}}s.
We first formulate the task (Section~\ref{sec:formulation}). Then, we present our reward design and action-level advantage estimation (Section~\ref{sec:reward}), and describe the training procedure (Section~\ref{sec:training}). 
Figure~\ref{fig:method} illustrates the overall framework.

\subsection{Task Formulation}\label{sec:formulation}

Given a repository and issue description $q$, our goal is to localize the code regions to be modified at three granularities, files $\hat{F}$, 
classes/modules $\hat{M}$, and functions $\hat{U}$, producing 
the prediction $\hat{y} = (\hat{F}, \hat{M}, \hat{U})$. 
The agent operates over an action space $\mathcal{A} = \{$\texttt{SearchCodeSnippet}, 
\texttt{TraverseGraph}, \texttt{SearchEntity}, \texttt{SearchSummary}, 
\texttt{ViewSummary}, \texttt{SearchCommit}$\}$ consisting of six tools that support keyword search, structural traversal, and semantic retrieval over the repository. Detailed explanations are provided in Appendix~\ref{app:tools}.

Following Section~\ref{sec:preliminary}, at each turn $t$ the policy $\pi_\theta$ generates a tool call $a_t \in \mathcal{A}$ from the current state $s_t$ and receives
an observation $o_t$, with $s_t$ encoding the issue $q$ and the interaction history
$(a_1, o_1, \ldots, a_{t-1}, o_{t-1})$.
Here, each action $a_t$ is defined as a tool name paired with its arguments, so that calls to the same tool with different arguments constitute distinct actions.
The process terminates with a \texttt{Finish}
action at turn $T$, yielding a trajectory $\tau = (q, a_1, o_1, \ldots, a_T, \hat{y})$.
Given a gold target $y = (F, M, U)$, we train $\pi_\theta$ to maximize
$\mathbb{E}_{\tau \sim \pi_\theta}[R(\tau; y)]$, where $R(\tau;y)$ decomposes into
turn-level rewards used to compute action-level advantages for fine-grained credit assignment.

\subsection{Reward Design and Action-Level Advantage Estimation}\label{sec:reward}
Localizing issue-relevant code regions requires sequential exploration, where each tool call shapes the search space for subsequent turns. However, trajectory-level signals alone face two limitations: they fail to capture partial successes where the agent uncovers the correct location but does not commit it to the final prediction, and Eq.~\eqref{eq:action_advantage} cannot be directly applied due to the lack of state equivalence in code exploration (Section~\ref{sec:preliminary}). We tackle these by designing a turn-level reward that captures partial successes (Section~\ref{sec:turn_reward}) and an action-level advantage estimation scheme over a code-specific state proxy (Section~\ref{sec:action_advantage}).

\subsubsection{Reward Design}
\label{sec:turn_reward}

In this section, we design rewards that operate at two complementary levels,
a \textit{terminal reward} at the final turn that evaluates the quality of the final prediction, and a \textit{turn-level reward} that provides immediate feedback whenever the agent newly discovers a gold location during exploration. Together they form the per-turn reward sequence $\{r_t\}_{t=1}^{T}$, from which the discounted return $G_t = \sum_{k=t}^{T} \gamma^{k-t} r_k$ is computed for action-level advantage estimation.

\paragraph{Terminal Reward.}
The terminal reward is computed at the final turn $T$ and reflects the quality of
the agent's final prediction. It consists of two complementary rewards:
$r_T = r^{\mathrm{F1}} + r^{\mathrm{DC}}$.

F1 reward $r^{\mathrm{F1}}$ evaluates final prediction accuracy as the sum of F1 scores at the file, module, and function levels:
\begin{equation}
\small
    r^{\mathrm{F1}} = \mathrm{F1}_F + \mathrm{F1}_M + \mathrm{F1}_U
    \label{eq:f1_reward}
\end{equation}

Discovery-Commit (DC) reward $r^{\mathrm{DC}}$ measures the fraction of explored gold locations that the agent commits to the final prediction. For each granularity level $g \in \{F, M, U\}$, we define the DC ratio as follows:
\begin{equation}
\small
    \mathrm{DC}_g = \frac{|E_g \cap Y_g \cap O_g|}{|E_g \cap Y_g|}
    \label{eq:dc_ratio}
\end{equation}
where $E_g$, $Y_g$, and $O_g$ denote the sets of explored, gold, and predicted locations at level $g$. Letting $\mathcal{I} = \{g \mid |E_g \cap Y_g| > 0\}$ be the set of valid levels, the DC reward is defined as follows:
\begin{equation}
\small
    r^{\mathrm{DC}} = \begin{cases} \dfrac{1}{|\mathcal{I}|} \sum_{g \in \mathcal{I}} \mathrm{DC}_g & \text{if } \mathcal{I} \neq \emptyset \\ 0 & \text{otherwise} \end{cases}
    \label{eq:dc_reward}
\end{equation}

Within the discounted return $G_t$, this terminal reward propagates the outcome quality back to earlier turns, encouraging actions that lead to both accurate final predictions and consistent commitment of discovered locations.

\paragraph{Turn-Level Reward.}
Relying solely on the terminal reward leaves no learning signal for failed trajectories ($G_t = 0$) even when gold locations were partially discovered, and underestimates the credit of early actions in successful trajectories due to discount-induced bias toward later turns. To tackle this, we additionally assign a turn-level reward that provides immediate feedback whenever the agent newly discovers a gold location.

Let $\delta^f_t$, $\delta^m_t$, $\delta^u_t$ denote the fraction of gold locations
newly discovered at turn $t$ at the file, module, and function levels. The turn-level reward is:
\begin{equation}
    r_t = \delta^f_t + \delta^m_t + \delta^u_t
    \label{eq:turn_reward}
\end{equation}
so that $G_t$ reflects per-turn discovery and preserves learning signals for partially correct trajectories that would otherwise receive no gradient.

\subsubsection{Action-Level Advantage Estimation}\label{sec:action_advantage}
Using $G_t$, we now estimate an action-level advantage that isolates the contribution of each individual action. As discussed in Section~\ref{sec:preliminary}, 
this requires identifying turns that took different actions under the same state $\tilde{s}$, and computing the advantage as the discounted return $G_t$ normalized against the mean return within such a group (Eq.~\eqref{eq:action_advantage}). The challenge is that code exploration lacks a natural notion of state equivalence: unlike game or web environments where the state changes through observable
transitions such as movement or page navigation, the repository remains fixed
throughout the trajectory, so a turn's state is determined solely by its
accumulated interaction history. Determining whether two
turns share the same state $\tilde{s}$ is therefore inherently difficult. We tackle this in two steps. We first define a state proxy for code exploration and group turns that share the same proxy. We then instantiate Eq.~\eqref{eq:action_advantage} on the resulting proxy-induced groups to obtain the action-level advantage.

\paragraph{State Proxy and Action-History Grouping.}

Treating two turns as sharing the same state $\tilde{s}$ based on their action alone is insufficient: the same action can carry different exploration contexts depending on when it is invoked within a trajectory. 

In our training data, approximately 79.2\% of turns involve a different preceding action, and the same action appears in an average of 3.29 distinct accumulated histories, indicating substantial contextual bias. We therefore define the state proxy based on the ``exploration history preceding'' the current turn. 
Specifically, the preceding $K$-action sequence at turn $t$ is $\phi_K(t) = \langle a_{t-K}, a_{t-K+1}, \ldots, a_{t-1} \rangle$ and turns sharing the same $\phi_K(t)$ are grouped into $\mathcal{G}^{(K)}_t$. Their mean return serves as a baseline for the average expected outcome at that exploration context. The current action $a_t$ is excluded from the group key, since including it would restrict comparisons to turns with the same action and cancel out the learning signal.

Note that grouping is performed per issue over rollouts of the same instance, and all exploration tools are read-only deterministic functions over a fixed repository, so an action with the same tool and arguments always returns the same observation. Turns sharing the same action history therefore reach their current turn through identical observations, ensuring that only comparable exploration contexts are grouped together.

\paragraph{Action-Level Advantage Estimation.}
Within a group $\mathcal{G}^{(K)}_t$ of turns sharing the same exploration history $\phi_K$, we instantiate Eq.~\eqref{eq:action_advantage} by normalizing the return $G_t$ against the group baseline and standard deviation:
\begin{equation}
\small
    \hat{A}^{(K)}_t = 
    \frac{G_t - \mathrm{mean}_{j \in \mathcal{G}^{(K)}_t} G_{t_j}}{\sigma_{\mathcal{G}^{(K)}_t} + \epsilon}.
    \label{eq:depth_advantage}
\end{equation}
Since the baseline is computed over turns that share the same exploration context but took alternative actions, $\hat{A}^{(K)}_t$ directly measures how much the current action $a_t$ outperformed its alternatives under the same context.

To avoid committing to a single depth and balance precision against stability, motivated by \citet{he2026hierarchy}, we combine advantages from multiple depths $K = 1, \ldots, H$, setting $\hat{A}^{(K)}_t = 0$ whenever the group size is one to suppress unreliable advantages. 
For turns whose groups are insufficient at every depth, we further introduce a trajectory-level anchor: the group-relative advantage $A(\tau_i)$ from Eq.~\eqref{eq:grpo} guarantees a minimum learning signal regardless of group availability.
Combining the two, the final turn-level advantage is defined as:
\begin{equation}
\small
    A_t = \omega_0 \cdot A(\tau_i) + \sum_{K=1}^{H} \omega_K \cdot \hat{A}^{(K)}_t,
    \label{eq:final_advantage}
\end{equation}
where $\omega_K \geq 0$ and $\sum_{K=0}^{H} \omega_K = 1$. The anchor term $\omega_0 \cdot A(\tau_i)$ ensures every turn receives at least a trajectory-level advantage, while each depth-$K$ term $\omega_K \cdot \hat{A}^{(K)}_t$ contributes action-level advantage only when sufficiently many turns share the same exploration context at that depth. 

\subsection{Two-Stage Training} \label{sec:training}

\paragraph{Teacher SFT Warm-up.}

Directly applying reinforcement learning to a pretrained LLM is challenging, as the initial policy lacks both the tool-use proficiency and the repository exploration patterns required to produce valid trajectories. 
To address this, we perform a SFT prior to reinforcement learning. 
Specifically, we leverage a strong teacher model to collect high-quality trajectories, filter for successful ones, and fine-tune the pretrained LLM on the resulting dataset to obtain the 
initial policy $\pi^{\mathrm{sft}}_\theta$.

\begin{table*}[!t]
\centering
\small
\setlength{\tabcolsep}{4pt}
\definecolor{diagbg}{RGB}{230, 230, 255}
\resizebox{\textwidth}{!}{%
\begin{tabular}{llccccccccc}
  \toprule
   &
    & \multicolumn{3}{c}{\textbf{File-level}}
    & \multicolumn{3}{c}{\textbf{Module-level}}
    & \multicolumn{3}{c}{\textbf{Function-level}} \\
  \cmidrule(lr){3-5}\cmidrule(lr){6-8}\cmidrule(lr){9-11}
  \textbf{Method} & \textbf{LLM} & $\mathrm{F1}$ & $P$ & $R$
    & $\mathrm{F1}$ & $P$ & $R$
    & $\mathrm{F1}$ & $P$ & $R$ \\
\midrule
\multicolumn{11}{c}{\textit{Closed-Source LLMs}} \\
\midrule
RepoSearcher      & Claude-3.7-Sonnet  & 32.30 & 20.24 & \textbf{89.24} & -- & -- & -- & 26.91 & 18.64 & \textbf{66.08} \\
\midrule
\multirow{3}{*}{RepoNavigator}
                  & GPT-5-Chat         & 58.88 & 61.87 & 58.17 & -- & -- & -- & 31.17 & 34.56 & 30.42 \\
                  & Claude-3.7-Sonnet  & 73.01 & 75.95 & 72.26 & -- & -- & -- & 31.72 & 34.43 & 31.03 \\
                  & Claude-Sonnet-4.5  & \underline{79.94} & \underline{81.92} & 80.68 & -- & -- & -- & 43.62 & 45.76 & 43.97 \\
\midrule
\multirow{2}{*}{OpenHands-Bash}
                  & GPT-5              & 3.20  & 3.20  & 3.20  & 2.60  & 2.60  & 2.60  & 2.60  & 2.60  & 2.60  \\
                  & Claude-Sonnet-4.5  & 0.80  & 0.80  & 0.80  & 0.40  & 0.40  & 0.40  & 0.40  & 0.40  & 0.40  \\
\midrule
\multirow{2}{*}{OpenHands-Bash$^{rem}$}
                  & GPT-5              & 78.18 & 79.25 & 80.80 & \underline{61.17} & \underline{62.23} & \underline{63.35} & \underline{54.79} & \underline{56.80} & 56.53 \\
                  & Claude-Sonnet-4.5  & \textbf{82.01} & \textbf{84.50} & \underline{82.86} & \textbf{67.19} & \textbf{70.11} & \textbf{67.47} & \textbf{61.78} & \textbf{65.42} & \underline{61.99} \\
\midrule
\multicolumn{11}{c}{\textit{Open-Source LLMs}} \\
\midrule
CoSIL             & \multirow{4}{*}{Qwen2.5-32B}
                                       & 30.77 & 19.34 & \underline{83.50} & -- & -- & -- & 22.11 & 14.85 & 55.38 \\
Agentless         &                    & 35.38 & 25.60 & 78.93 & -- & -- & -- & 27.33 & 24.07 & 40.97 \\
LocAgent          &                    & 44.18 & 34.18 & 79.39 & -- & -- & -- & 21.48 & 16.29 & 46.79 \\
OrcaLoca          &                    & 58.11 & 59.51 & 59.57 & -- & -- & -- & 28.72 & 25.59 & 39.14 \\
\midrule
\multirow{2}{*}{RepoSearcher}
                  & Qwen2.5-7B  & 30.09 & 18.80 & 83.11 & -- & -- & -- & 25.57 & 17.68 & \underline{62.38} \\
                  & Qwen2.5-32B & 32.25 & 20.24 & \textbf{88.59} & -- & -- & -- & 28.03 & 19.36 & \textbf{68.55} \\
\midrule
\multirow{3}{*}{RepoNavigator}
                  & Qwen2.5-7B   & 51.63 & 53.83 & 50.62 & -- & -- & -- & 27.49 & 30.34 & 26.69 \\
                  & Qwen2.5-14B  & 58.90 & 58.97 & 61.60 & -- & -- & -- & 29.23 & 30.08 & 31.02 \\
                  & Qwen2.5-32B  & 67.75 & 70.76 & 67.29 & -- & -- & -- & 34.09 & 37.19 & 33.71 \\
\midrule
\multirow{6}{*}{OpenHands-Bash}
                  & Qwen3-1.7B               & 2.40  & 2.09  & 3.60  & 0.37  & 0.32  & 0.60  & 0.34  & 0.32  & 0.50  \\
                  & Qwen3-4B-Instruct        & 49.73 & 49.69 & 53.34 & 19.32 & 19.86 & 20.15 & 13.27 & 14.17 & 13.74 \\
                  & Qwen3-14B                & 43.13 & 36.49 & 71.20 & 22.86 & 20.40 & 33.04 & 16.08 & 14.51 & 23.58 \\
                  & Qwen3-32B (Thinking)     & 62.91 & 59.87 & 73.63 & 34.69 & 33.85 & 39.46 & 23.99 & 24.22 & 26.86 \\
                  & \textsc{CodeScout}-4B    & 68.52 & \textbf{71.53} & 67.74 & 45.97 & \underline{49.70} & 44.97 & 36.78 & \underline{40.71} & 35.72 \\
                  & \textsc{CodeScout}-14B   & 68.57 & \underline{71.00} & 68.69 & \textbf{50.88} & \textbf{53.71} & 50.88 & \underline{40.32} & \textbf{43.74} & 40.27 \\
\midrule
\cellcolor{diagbg} & \cellcolor{diagbg}Qwen3-4B-Instruct & \cellcolor{diagbg}\underline{72.64} & \cellcolor{diagbg}70.88 & \cellcolor{diagbg}76.76 & \cellcolor{diagbg}46.94 & \cellcolor{diagbg}42.25 & \cellcolor{diagbg}\underline{59.55} & \cellcolor{diagbg}38.59 & \cellcolor{diagbg}35.50 & \cellcolor{diagbg}48.98 \\
\cellcolor{diagbg}\multirow{-2}{*}{\textsc{$\text{A}^2$Agent}} & \cellcolor{diagbg}Qwen3-8B & \cellcolor{diagbg}\textbf{73.31} & \cellcolor{diagbg}70.14 & \cellcolor{diagbg}80.23 & \cellcolor{diagbg}\underline{47.98} & \cellcolor{diagbg}42.56 & \cellcolor{diagbg}\textbf{63.42} & \cellcolor{diagbg}\textbf{41.01} & \cellcolor{diagbg}37.15 & \cellcolor{diagbg}54.44 \\
\bottomrule
\end{tabular}%
}
\vspace{-0.2cm}
\caption{Comparison of different methods on file-level, module-level, and
  function-level on SWE-Bench Verified using three evaluation metrics, F1 score
  ($\mathrm{F1}$), Precision ($P$), and Recall ($R$). The best result per metric is
  \textbf{bold-faced} and the second best is \underline{underlined} within each
  block (closed- and open-source LLMs).}
\label{tab:main}
\end{table*}
\begin{table*}[t]
\centering
\small
\setlength{\tabcolsep}{4pt}

\definecolor{diagbg}{RGB}{230, 230, 255}
\resizebox{\textwidth}{!}{%
\begin{tabular}{llccccccccc}
  \toprule
   &
    & \multicolumn{3}{c}{\textbf{File-level}}
    & \multicolumn{3}{c}{\textbf{Module-level}}
    & \multicolumn{3}{c}{\textbf{Function-level}} \\
  \cmidrule(lr){3-5}\cmidrule(lr){6-8}\cmidrule(lr){9-11}
  \textbf{Method} & \textbf{LLM} & $\mathrm{F1}$ & $P$ & $R$
    & $\mathrm{F1}$ & $P$ & $R$
    & $\mathrm{F1}$ & $P$ & $R$ \\
\midrule
\multicolumn{11}{c}{\textit{Closed-Source LLMs}} \\
\midrule
\multirow{2}{*}{OpenHands-Bash}
                  & GPT-5              & 0.00  & 0.00  & 0.00  & 0.00  & 0.00  & 0.00  & 0.00  & 0.00  & 0.00  \\
                  & Claude-Sonnet-4.5  & 0.00  & 0.00  & 0.00  & 0.00  & 0.00  & 0.00  & 0.00  & 0.00  & 0.00  \\
\midrule
\multirow{2}{*}{OpenHands-Bash$^{rem}$}
                  & GPT-5              & \underline{61.18} & \underline{69.10} & \underline{62.06} & \underline{42.20} & \underline{52.86} & \underline{39.63} & \underline{35.86} & \underline{48.91} & \underline{32.65} \\
                  & Claude-Sonnet-4.5  & \textbf{64.75} & \textbf{75.39} & \textbf{64.49} & \textbf{48.54} & \textbf{61.75} & \textbf{45.39} & \textbf{42.26} & \textbf{58.14} & \textbf{38.06} \\
\midrule
\multicolumn{11}{c}{\textit{Open-Source LLMs}} \\
\midrule
RepoSearcher      & \multirow{4}{*}{Qwen2.5-32B}
                                       & 3.81  & 2.52  & 9.00  & --    & --    & --    & 2.31  & 2.46  & 2.52  \\
LocAgent          &                    & 19.77 & 0.38  & 25.73 & --    & --    & --    & 4.30  & 0.17  & 8.72  \\
Agentless         &                    & 20.07 & 13.89 & 43.07 & --    & --    & --    & 7.98  & 7.31  & 11.08 \\
CoSIL             &                    & 20.95 & 14.03 & 48.87 & --    & --    & --    & 7.67  & 6.00  & 14.03 \\
\midrule
\multirow{3}{*}{RepoNavigator}
                  & Qwen2.5-7B   & 39.74 & 48.13 & 36.36 & --    & --    & --    & 14.29 & 21.26 & 12.33 \\
                  & Qwen2.5-14B  & 49.72 & 58.64 & 46.85 & --    & --    & --    & 18.06 & 25.25 & 16.05 \\
                  & Qwen2.5-32B  & 57.57 & 68.69 & 53.49 & --    & --    & --    & 20.72 & 29.44 & 18.13 \\
\midrule
\multirow{6}{*}{OpenHands-Bash}
                  & Qwen3-1.7B               & 0.73  & 0.72  & 1.13  & 0.00  & 0.00  & 0.00  & 0.00  & 0.00  & 0.00  \\
                  & Qwen3-4B-Instruct        & 36.96 & 44.42 & 35.59 & 11.78 & 17.46 & 10.19 & 8.12  & 12.16 & 7.01  \\
                  & Qwen3-14B                & 30.08 & 28.48 & 48.22 & 11.87 & 13.82 & 14.21 & 8.20  & 9.97  & 9.92  \\
                  & Qwen3-32B (Thinking)     & 46.85 & 49.65 & \underline{54.55} & 21.94 & 27.18 & 22.62 & 12.31 & 17.82 & 11.93 \\
                  & \textsc{CodeScout}-4B    & 51.77 & 68.98 & 46.16 & 36.97 & \textbf{56.05} & 31.13 & 29.03 & \textbf{48.65} & 23.73 \\
                  & \textsc{CodeScout}-14B   & 53.63 & 68.81 & 48.81 & 37.13 & \underline{53.80} & 32.02 & 28.74 & \underline{46.09} & 23.76 \\
\midrule
\cellcolor{diagbg} & \cellcolor{diagbg}Qwen3-4B-Instruct & \cellcolor{diagbg}\underline{58.19} & \cellcolor{diagbg}\underline{73.34} & \cellcolor{diagbg}53.83 & \cellcolor{diagbg}\underline{38.38} & \cellcolor{diagbg}45.18 & \cellcolor{diagbg}\underline{40.00} & \cellcolor{diagbg}\underline{29.23} & \cellcolor{diagbg}38.29 & \cellcolor{diagbg}\underline{28.85} \\
\cellcolor{diagbg}\multirow{-2}{*}{\textsc{$\text{A}^2$Agent}} & \cellcolor{diagbg}Qwen3-8B & \cellcolor{diagbg}\textbf{60.05} & \cellcolor{diagbg}\textbf{76.54} & \cellcolor{diagbg}\textbf{56.16} & \cellcolor{diagbg}\textbf{39.07} & \cellcolor{diagbg}45.51 & \cellcolor{diagbg}\textbf{42.37} & \cellcolor{diagbg}\textbf{30.60} & \cellcolor{diagbg}39.16 & \cellcolor{diagbg}\textbf{30.91} \\
\bottomrule
\end{tabular}%
}
\vspace{-0.2cm}
\caption{Comparison of different methods on file-level, module-level, and
  function-level on SWE-Bench Pro using three evaluation metrics, F1 score ($\mathrm{F1}$),
  Precision ($P$), and Recall ($R$). The best result per metric is
  \textbf{bold-faced} and the second best is \underline{underlined} within each
  block (closed- and open-source LLMs).}
\label{tab:main-pro}
\end{table*}

\paragraph{Action-Aware Reinforcement Learning.}
Initialized from $\pi^{\mathrm{sft}}_\theta$, we perform action-aware reinforcement learning using the advantage $A_t$ defined in Section~\ref{sec:reward} to improve code localization. We first collect $N$ trajectories per issue from the training set. To reinforce actions at high-advantage turns and down-weight those at low-advantage turns, we define an advantage-weighted maximum-likelihood objective \citep{peng2019advantage} that assigns token weights proportional to $A_t$. The transformation from $A_t$ to a positive token weight is as follows:

\begingroup
\small
\begin{align}
    \tilde{A}_t &= \text{clip}(A_t, -\kappa, \kappa), \nonumber \\
    w_t &= \text{clip}\left(\exp(\beta \tilde{A}_t), w_\text{min}, w_\text{max}\right)
    \label{eq:weight}
\end{align}
\endgroup
The inner clip bounds outlier advantages and the outer clip keeps the resulting positive weight within a stable range.
The training objective is defined as follows, where the token weight $w_i := w_{t_i}$ assigns each token $i$ the weight of its corresponding turn $t_i$, and $\ell_i(\theta) = -\log \pi_\theta(y_i \mid y_{<i}, \tau_{<t_i})$ denotes the per-token cross-entropy of policy $\pi_\theta$:
\begin{equation}
\small
    \mathcal{L}(\theta) = \mathbb{E}_{\tau \sim \mathcal{D}} \left[ \frac{1}{|V(\tau)|} \sum_{i \in V(\tau)} w_i \cdot \ell_i(\theta) \right]
    \label{eq:loss}
\end{equation}
where $\mathcal{D}$ is the set of collected trajectories and
$V(\tau)$ denotes the index set of all assistant tokens in trajectory
$\tau$. The overall training algorithm is summarized in Appendix~\ref{app:algorithms}.

\section{Experiments}
\label{sec:experiments}

\subsection{Experimental Setup}
\label{sec:exp-setup}

\paragraph{Datasets.}
We construct 2,438 training instances from SWE-Gym~\cite{pan2024training}, 
and evaluate on two held-out benchmarks: SWE-Bench Verified~\cite{jimenez2024swebench} 
(500 human-verified) and the more challenging SWE-Bench Pro~\cite{deng2025swe} 
(266 instances) for generalization. We exclude any repository-level overlap between training and evaluation sets to prevent contamination. Further details are provided in Appendix~\ref{app:benchmark}.

\paragraph{Baselines.}
We compare our method against several representative baselines using both open- and closed-source LLMs, including 
CoSIL~\cite{jiang2025cosil}, Agentless~\cite{xia2024agentless}, 
LocAgent~\cite{chen2025locagent}, 
OrcaLoca~\cite{yu2025orcaloca}, 
RepoSearcher~\cite{ma2025tool}, 
RepoNavigator~\cite{zhang2025one}, and 
CodeScout~\cite{sutawika2026codescout}.
All baseline results are taken from the respective papers or subsequent work. To provide closed-source upper bounds, we additionally evaluate GPT-5 and Claude-Sonnet-4.5 via the OpenHands~\cite{wang2025openhands} scaffold. Detailed descriptions are provided in Appendix~\ref{app:baselines}.
 
\paragraph{Metrics.}

Following~\citet{sutawika2026codescout}, 
we predict a variable number of locations per issue rather than a fixed-length candidate list, allowing the model to adaptively determine the proper scope of each prediction.
Under this setting, our primary evaluation metric is the instance-level average \textit{F1 score} between predicted and ground-truth locations across three granularities of file, module, and function.
We additionally report \textit{precision} and \textit{recall} at each granularity to enable fair comparison with baselines that predict differing numbers of locations. The formal definition of the evaluation protocol is provided in Appendix~\ref{app:eval_protocol}.

\paragraph{Implementation Details.}
We train and evaluate with Qwen3-4B-Instruct and Qwen3-8B as base models, fine-tuned with LoRA (rank${=}16$) for 3 epochs. For RL, we run 2 epochs with a batch size of 16, sampling $N{=}8$ trajectories per issue and setting the history depth to $H{=}2$. Further implementation details are provided in Appendix~\ref{app:impl}.

\subsection{Main Results}
\label{sec:main-results}

As reported in Table~\ref{tab:main}, \textsc{$\text{A}^2$Agent} consistently outperforms all open-source baselines in average F1 over SWE-Bench Verified, surpassing the SOTA baseline (CodeScout-14B) by 1.58\%. Notably, our 4B model even outperforms baselines up to 8$\times$ larger (32B-scale models). Here, OpenHands-Bash$^{rem}$ denotes an OpenHands-Bash variant with patch-editing removed, restricting it to localization.
Trajectory-level training methods such as RepoNavigator and CodeScout assign a single advantage to every turn, capturing only the overall trajectory quality and leaving each individual tool call's contribution unattributed. In contrast, \textsc{$\text{A}^2$Agent} rewards each call by comparing it against alternative actions taken under the same exploration history, converting a sparse reward into a dense action-level reward. This enables even a small model to learn targeted search over redundant broad search and to commit more of the gold locations it uncovers.

Furthermore, as shown in Table~\ref{tab:main-pro}, \textsc{$\text{A}^2$Agent} also consistently outperforms all open-source baselines across all three granularities on SWE-Bench Pro, demonstrating robust generalization to diverse repositories and varying levels of task difficulty. We also provide qualitative results in Appendix~\ref{app:case_study}.

\begin{table}[t]
\centering
\begin{adjustbox}{max width=\linewidth}
\begin{tabular}{lccccc}
\toprule
\textbf{Method} & \textbf{\makecell{History\\Grouping}} & \textbf{$r^{\mathrm{DC}}$} & \textbf{\makecell{File\\F1}} & \textbf{\makecell{Module\\F1}} & \textbf{\makecell{Function\\F1}} \\ \midrule
\multirow{4}{*}{\textsc{$\text{A}^2$Agent}} & \cmark & \cmark & \textbf{73.31} & \textbf{47.98} & \textbf{41.01} \\ \cmidrule{2-6}
 & \cmark & \xmark & \makecell{71.73 \\ \textcolor{red}{\small(-1.58)}} & \makecell{46.42 \\ \textcolor{red}{\small(-1.56)}} & \makecell{39.55 \\ \textcolor{red}{\small(-1.46)}} \\
 & \xmark & \cmark & \makecell{72.19 \\ \textcolor{red}{\small(-1.12)}} & \makecell{47.10 \\ \textcolor{red}{\small(-0.88)}} & \makecell{40.19 \\ \textcolor{red}{\small(-0.82)}} \\
 & \xmark & \xmark & \makecell{71.01 \\ \textcolor{red}{\small(-2.30)}} & \makecell{45.63 \\ \textcolor{red}{\small(-2.35)}} & \makecell{38.72 \\ \textcolor{red}{\small(-2.29)}} \\ \midrule
SFT & - & - & \makecell{66.20 \\ \textcolor{red}{\small(-7.11)}} & \makecell{41.15 \\ \textcolor{red}{\small(-6.83)}} & \makecell{35.41 \\ \textcolor{red}{\small(-5.60)}} \\
GRPO & - & - & \makecell{67.47 \\ \textcolor{red}{\small(-5.84)}} & \makecell{42.33 \\ \textcolor{red}{\small(-5.65)}} & \makecell{34.28 \\ \textcolor{red}{\small(-6.73)}} \\ \bottomrule
\end{tabular}
\end{adjustbox}
\caption{Ablation study on key components of \textsc{$\text{A}^2$Agent}. All ablations are conducted under the Qwen3-8B backbone on SWE-Bench Verified. Values in parentheses indicate the drop in percentage points relative to ours.
}
\vspace{-0.5cm}
\label{tab:ablation}
\end{table}

\subsection{Ablation Study}
\label{sec:ablation}

In Table~\ref{tab:ablation}, we conduct an ablation study to verify the effectiveness of our method and the contribution of each proposed component, comparing ours against SFT and GRPO baselines while ablating its two key components, History Grouping ($H{=}2$) and Discovery-Commit reward ($r^{\mathrm{DC}}$).

First, SFT generalizes poorly to unseen codebases (7.11\%p drop in file-level F1), while GRPO assigns the same advantage to every turn, leaving individual tool call contributions unattributed and treating partially successful trajectories identically to complete failures.
In contrast, even the simplest variant of \textsc{$\text{A}^2$Agent} (both components removed) already outperforms GRPO by 3.54\%p, demonstrating that turn-level reward alone provides substantially richer learning signals than trajectory-level supervision.

Second, both components contribute complementarily. History Grouping and $r^{\mathrm{DC}}$ each contribute independently and combine for the best result, as they target distinct failure modes: History Grouping refines action-level credit by comparing turns under matched exploration contexts, while $r^{\mathrm{DC}}$ explicitly rewards the agent for committing the gold elements it observes during exploration.

Finally, these contributions hold consistently across all three granularities. Their effect does not diminish at finer levels. While the file level has a small candidate space that even coarse exploration can reach, the function level relies more heavily on committing discovered locations and isolating per-action credit, which is precisely where our components operate.

\begin{figure}[t]
  \centering
  \includegraphics[width=\linewidth]{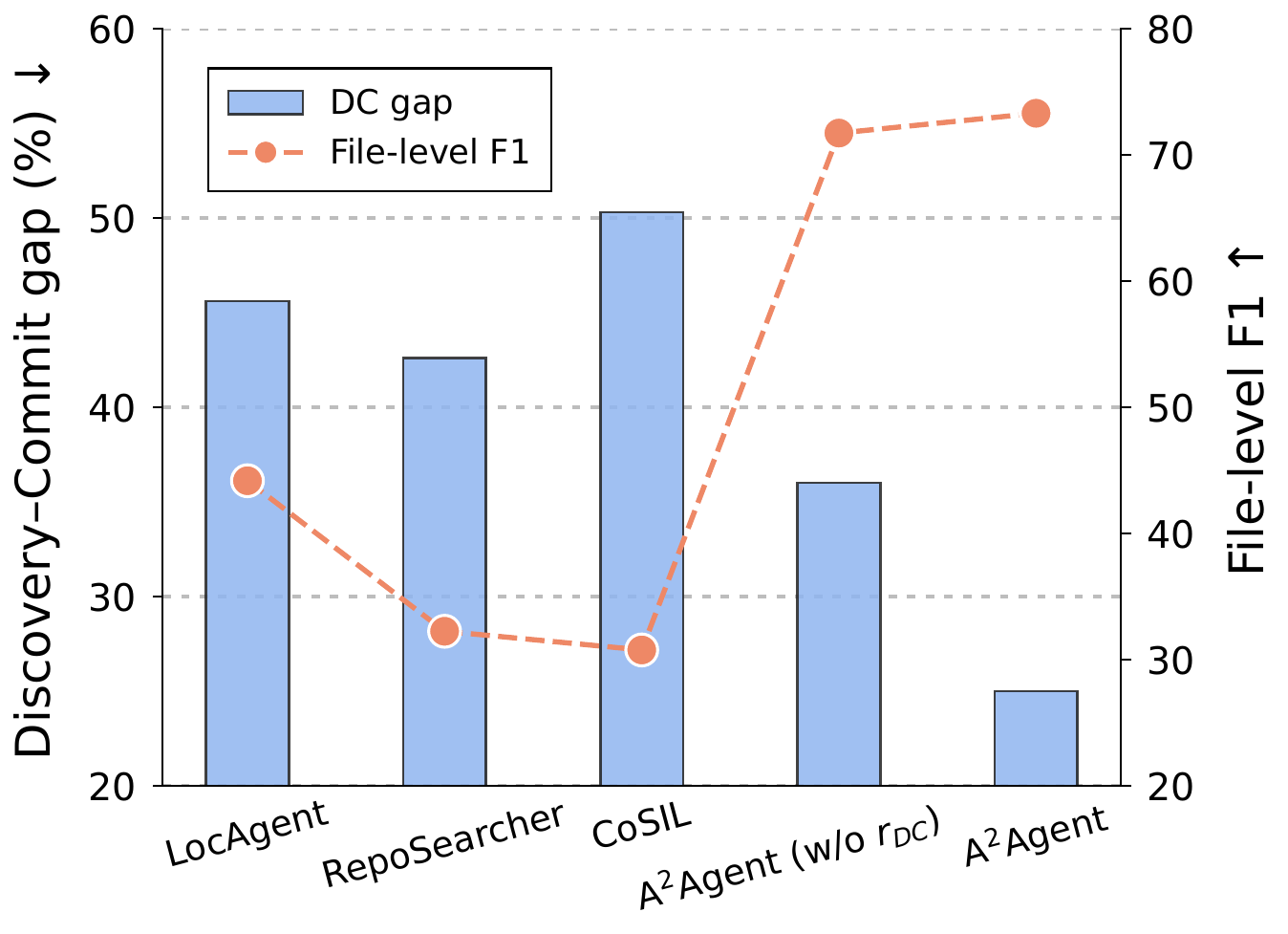}
  \vspace{-0.5cm}
  \caption{
    Comparison of Discovery-Commit gap and file-level F1 under Qwen3-8B backbone. $r^{\mathrm{DC}}$ markedly reduces the gap while keeping the highest F1.
  }
  \label{fig:dc_gap}
\end{figure}

\begin{figure}[t]
    \centering
    \includegraphics[width=\linewidth]{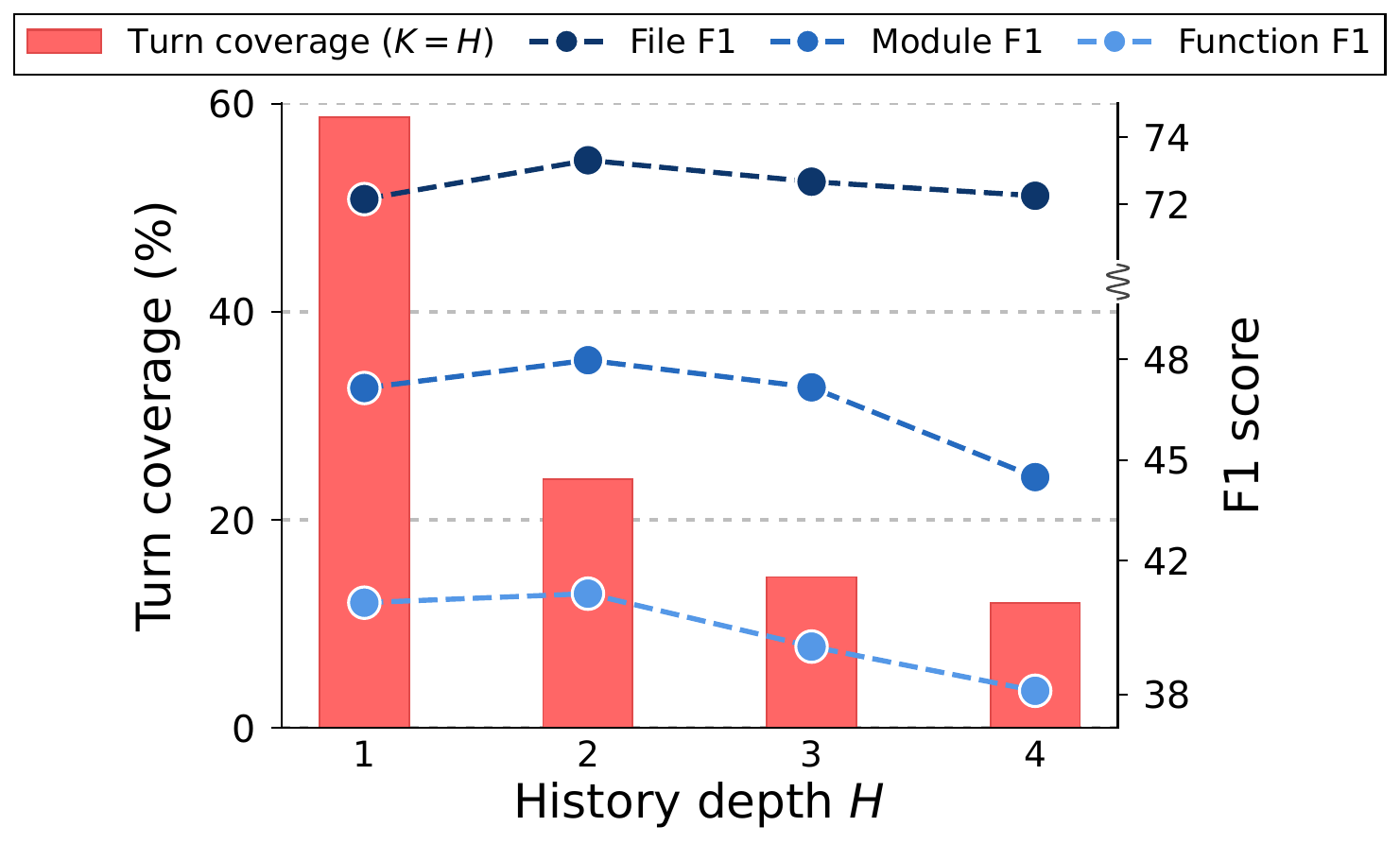}
    \caption{Effect of history depth $H$ on F1 and turn coverage under Qwen3-8B backbone, the fraction of turns in groups of size $\geq 2$ at depth $K{=}H$. F1 peaks at $H{=}2$ and declines as coverage shrinks with larger $H$.}
    \label{fig:depth_analysis}
\end{figure}

\subsection{Further Analysis of \textsc{$\text{A}^2$Agent}}
\label{sec:effectiveness}

To further analyze the effectiveness of our method, we measure the \textit{Discovery-Commit gap} across baselines in Figure~\ref{fig:dc_gap}, defined as the fraction of failures in which a gold location was discovered during exploration but not committed to the final prediction. This gap accounts for over 40\% of failures across all baselines, revealing uncommitted discovery as a prevalent failure mode under trajectory-level supervision. 
In contrast, ours with $r^{\mathrm{DC}}$ substantially reduces this gap while simultaneously improving file-level F1, confirming that $r^{\mathrm{DC}}$ converts explored gold locations into correct predictions rather than trading localization accuracy for commitment.

To assess the effect of history depth $H$, we measure F1 and turn coverage (the fraction of turns in groups of size $\geq 2$ at depth $K{=}H$) across $H$ values in Figure~\ref{fig:depth_analysis}.
Increasing $H$ trades coverage for precision: smaller $H$ groups more turns under a coarser context, while larger $H$ matches turns under a finer context but leaves fewer turns grouped. Beyond $H{=}2$, coverage drops steeply and groups become too small for reliable credit estimation. The added deep-level terms thus contribute more noise than signal to Eq.~\eqref{eq:final_advantage}, causing F1 to decline. We therefore set $H{=}2$, which best balances grouping precision and coverage.

\subsection{Inference Efficiency Analysis} \label{sec:efficiency}
 
\begin{table}[t]
\centering
\begin{adjustbox}{max width=\linewidth}
\begin{tabular}{llcc}
\toprule
\textbf{Type} & \textbf{Method} & \textbf{\makecell{Avg. \\ \#Turns ($\downarrow$)}} & \textbf{\makecell{Avg. \\ Time ($\downarrow$)}} \\
\midrule
Procedure-based & Agentless & 2.77 & 6.41 \\
\midrule
\multirow{4}{*}{Agent-based}
                & CoSIL        & 17.55 & 77.57 \\
                & LocAgent     & \textbf{6.20}  & 86.23 \\
                & RepoSearcher & 12.47 & 42.58 \\
                & \textsc{$\text{A}^2$Agent}        & 11.72 & \textbf{18.08} \\
\bottomrule
\end{tabular}
\end{adjustbox}
\caption{Inference efficiency comparison on SWE-Bench Verified, where all methods use the Qwen3-4B-Instruct backbone for fair comparison. \textbf{Avg.\ \#Turns}: average number of tool calls per instance. \textbf{Avg.\ Time}: average inference time per instance.}
\label{tab:efficiency}
\end{table}

To assess the practical cost of \textsc{$\text{A}^2$Agent}, we report inference efficiency in Table~\ref{tab:efficiency}. Among agent-based methods, \textsc{$\text{A}^2$Agent} attains the lowest inference time, which stems not from fewer tool calls but from more concise observations. Without action-level credit, baselines repeatedly invoke broad exploration tools that contribute little to localization. 
In contrast, \textsc{$\text{A}^2$Agent} learns to shift from broad exploration to targeted search, reducing average observation length by approximately 40\% compared to baselines. Since observations accumulate in the prefill context of later turns, this gap compounds into a substantial speedup.

\subsection{Influence on Issue Resolution}
\label{sec:downstream}

To verify that our localization gains carry over to end-to-end issue resolution, we conduct a downstream experiment that measures resolve rate directly, as reported in Table~\ref{tab:downstream}.
We isolate the effect of localization by fixing the repair back-end to the Agentless repair pipeline. Every method feeds its localization output into this identical pipeline, performing localization with Qwen3-4B-Instruct and patch generation with Qwen3-30B-A3B-Instruct, so any difference in resolve rate reflects localization quality alone. Oracle feeds the gold locations directly into the repair pipeline and thus marks the upper bound reachable with perfect localization.
Resolve rate rises together with localization quality across all methods, and \textsc{$\text{A}^2$Agent} reaches the highest resolve rate among all baselines while substantially narrowing the gap to the Oracle upper bound. This confirms that the F1 gains from \textsc{$\text{A}^2$Agent} translate into a clear improvement in end-to-end resolve rate, consistent with the widely recognized view that localization quality directly bounds final resolution performance \citep{jimenez2024swebench, xia2024agentless}.

\begin{table}[t]
\centering
\begin{adjustbox}{max width=\linewidth}
\begin{tabular}{lccc}
\toprule
\textbf{Method} & \textbf{File F1} & \textbf{Function F1} & \textbf{Resolved (\%)} \\ \midrule
Oracle & - & - & 45.93 \\ \midrule
RepoSearcher & 29.33 & 18.02 & 26.82 \\
CoSIL & 32.70 & 19.37 & 27.64 \\
Agentless & 44.70 & 20.54 & 29.07 \\
LocAgent & 49.14 & 25.82 & 31.78 \\
\textsc{$\text{A}^2$Agent} & \textbf{72.64} & \textbf{38.59} & \textbf{38.07} \\ \bottomrule
\end{tabular}
\end{adjustbox}
\caption{Downstream issue resolution results on SWE-Bench Verified. All localization methods use the Qwen3-4B-Instruct backbone, and their outputs are fed into the same repair pipeline. Oracle denotes providing the gold locations directly to the repair pipeline.}
\vspace{-0.4cm}
\label{tab:downstream}
\end{table}
\section{Conclusion}
\label{sec:conclusion}
We proposed \textsc{$\text{A}^2$Agent}, an action-aware reinforcement learning method for repository-level code localization agents. By estimating action-level advantages and incorporating a Discovery-Commit reward, the agent learns which tool calls advance localization and reliably commits discovered gold locations. Extensive experiments demonstrated that \textsc{$\text{A}^2$Agent} achieved strong performance, outperforming baselines up to 8$\times$ larger. Requiring no external API access or Docker setup, \textsc{$\text{A}^2$Agent} can be deployed entirely on-premises, making it practical where source code privacy is required.

\section*{Limitations}
\label{sec:limitation}

Owing to computational resource constraints, our experiments were conducted on small open-source models rather than larger-scale ones. Even so, our method achieved strong performance at this scale. Moreover, although our method was also effective for downstream issue resolution as shown in Table~\ref{tab:downstream}, further optimization for the downstream task remains a promising direction for our future work.

\section*{Acknowledgements}
This work was supported by Institute of Information \& communications Technology Planning \& Evaluation (IITP) grant funded by the Korea government(MSIT) (NO.RS-2025-02218768, 20\%; No.RS-2025-25442569, 20\%; RS-2019-II190421, 20\%; IITP-2026-RS-2024-00437633, 20\%), and the National Research Foundation of Korea(NRF) funded by the Ministry of Education (RS-2025-25433088, 20\%).

\bibliography{custom}

\clearpage
\appendix
\appendix
 
\section{Tool Specifications}
\label{app:tools}
 
In this section, we describe the six repository tools comprising the action space $\mathcal{A}$ introduced in Section~\ref{sec:formulation}. The tools are built
on a repository graph index, a BM25-based code search index, a semantic
file summary index, and a commit history index. We directly adopt the tool
infrastructure from prior work~\cite{chen2025locagent, wang2025improving},
ensuring that any performance gains observed in our experiments are
attributable solely to our learning signal design rather than differences
in tool design.
 
\paragraph{\texttt{TraverseGraph(start, depth, edges)}.}
Returns a subgraph rooted at the given entity, traversing up to \texttt{depth} hops along the specified dependency types (\texttt{contains}, \texttt{imports}, \texttt{invokes}, \texttt{inherits}). Serves as the primary entry point for understanding the module hierarchy and dependency structure of an unfamiliar
repository.
 
\paragraph{\texttt{SearchCodeSnippet(query)}.}
Searches the entire codebase via a BM25 index using keywords or identifiers,
returning the file path, line range, and a snippet of the top matching
results. Most effective when exact identifiers can be extracted from the
issue description and used to narrow down candidates for subsequent
\texttt{SearchEntity} calls.
 
\paragraph{\texttt{SearchEntity(entity)}.}
Returns a path of the form \texttt{file\_path}, \texttt{file\_path:Class},
or \texttt{file\_path:Class.method}, and returns the raw source code of the
specified entity. Used to directly inspect the logic of a narrowed candidate
location and confirm a function-level answer.
 
\paragraph{\texttt{SearchSummary(query)}.}
Searches file summaries generated by an LLM by semantic similarity against
a natural language query, returning the top file paths and their summaries.
Useful for narrowing the file candidate space when the issue is described
in domain-level language without explicit identifiers.
 
\paragraph{\texttt{ViewSummary(file)}.}
Returns a file summary generated by an LLM, covering its key
responsibilities, major classes and functions, and dependencies. Enables
rapid understanding of a file's role without reading its full source.
 
\paragraph{\texttt{SearchCommit(query)}.}
Returns the SHA, message, date, and list of changed files for past commits
matching the given query. Provides context from historically similar changes and co-modified file dependencies, supplementing static analysis with dynamic repository evolution signals.

  \begin{table}[t]
  \centering
  \small
  \setlength{\tabcolsep}{4pt}
  \renewcommand{\arraystretch}{1.1}
  \resizebox{\columnwidth}{!}{%
  \begin{tabular}{l c}
  \toprule
  \textbf{Hyperparameter} & \textbf{Value} \\
  \midrule
  Rollouts per issue $N$                        & 8 \\
  Max tool calls per trajectory                 & 20 \\
  History depth $H$                             & 2 \\
  Discount factor $\gamma$                      & 0.9 \\
  Batch size                                    & 16 \\
  Depth weights $\omega_K$                      & $\propto K{+}1$ \\
  Advantage clip $\kappa$                       & 5.0 \\
  Token-weight temperature $\beta$              & 1.5 \\
  Token-weight bounds $w_{\min}$, $w_{\max}$    & 0.1, 3.0 \\
  LoRA rank, $\alpha$                           & 16, 16 \\
  Learning rate (SFT)                           & $2{\times}10^{-4}$ \\
  Learning rate (advantage-weighted RL)         & $5{\times}10^{-5}$ \\
  Learning rate schedule                        & Linear w/ warmup \\
  Optimizer                                     & AdamW \\
  Weight decay                                  & 0.01 \\
  \bottomrule
  \end{tabular}%
  }
  \caption{Hyperparameters used for \textsc{$\text{A}^2$Agent}.}
  \label{tab:hparams}
  \end{table}

\begin{table*}[t]
\centering
\small
\setlength{\tabcolsep}{3pt}
\renewcommand{\arraystretch}{1.15}
\resizebox{\textwidth}{!}{%
\begin{tabular}{l l p{3.6cm} p{3.8cm}}
\toprule
\textbf{Method} & \textbf{Tool} & \textbf{Input} & \textbf{Output} \\
\midrule
Agentless~\cite{xia2024agentless}        & none (fixed pipeline)        & problem description, file tree                  & ranked file/class/function list \\
\midrule
CoSIL~\cite{jiang2025cosil}              & symbol-trace                 & symbol name                                     & call-graph slice rooted at symbol \\
\midrule
\multirow{5}{*}{OrcaLoca~\cite{yu2025orcaloca}}
                                          & \texttt{search\_file\_contents}   & file path                                  & file body or skeleton \\
                                          & \texttt{search\_class}            & class name                                 & class definition \\
                                          & \texttt{search\_method\_in\_class}& class, method name                         & method body \\
                                          & \texttt{search\_callable}         & callable name                              & callable code snippet \\
                                          & \texttt{search\_source\_code}     & code string                                & matching snippets \\
\midrule
\multirow{4}{*}{LocAgent~\cite{chen2025locagent}}
                                          & \texttt{SearchEntity}        & entity name                                     & matching entity records \\
                                          & \texttt{TraverseGraph}       & start node, direction, depth                    & graph slice \\
                                          & \texttt{RetrieveEntity}      & entity ID                                       & entity code block \\
                                          & \texttt{FuzzySearch}         & fuzzy query                                     & candidate entities \\
\midrule
\multirow{4}{*}{RepoSearcher~\cite{ma2025tool}}
                                          & \texttt{GetRepoStructure}    & repository root                                 & directory tree \\
                                          & \texttt{SearchClass}         & class name                                      & class body \\
                                          & \texttt{SearchFunction}      & function name                                   & function body \\
                                          & \texttt{SearchClassMethod}   & class.method name                               & method body \\
\midrule
RepoNavigator~\cite{zhang2025one} & \texttt{Jump}              & symbol name in a file                           & symbol definition \\
\midrule
\multirow{2}{*}{CodeScout~\cite{sutawika2026codescout}}
                                          & \texttt{Terminal}            & Unix shell command                              & command stdout/stderr \\
                                          & \texttt{LocalizationFinish}  & \{file, module, function\} sets                 & final structured answer \\
\midrule
\multirow{6}{*}{\textsc{$\text{A}^2$Agent}}                     & \texttt{SearchEntity}        & entity name                                     & entity source code \\
                                          & \texttt{TraverseGraph}       & start node, direction, depth                    & graph slice \\
                                          & \texttt{SearchCodeSnippet}   & keyword or identifier                           & matching code snippets \\
                                          & \texttt{SearchSummary}       & natural language query                          & file paths and summaries \\
                                          & \texttt{ViewSummary}         & file path                                       & natural language file summary \\
                                          & \texttt{SearchCommit}        & keyword                                         & matching commit history \\
\bottomrule
\end{tabular}%
}
\caption{Descriptions of the tool sets of code localization agents. \textsc{$\text{A}^2$Agent} adopts the same tool infrastructure as existing methods, focusing on training
methodology rather than novel tool design.}
\label{tab:tool-comparison}
\end{table*}

\begin{algorithm}[t]
  \caption{Training Procedure of \textsc{$\text{A}^2$Agent}}
  \label{alg:training}
  \small    
  \begin{algorithmic}[1]
  \Require SFT policy $\pi^{\mathrm{sft}}_\theta$, training queries $\mathcal{Q}$, rollouts per
  issue $N$, history depth $H$, discount factor $\gamma$, advantage weights $\omega_{0:H}$
  \State $\theta \gets \theta^{\mathrm{sft}}$
  \Repeat
      \State $\mathcal{D} \gets \emptyset$
      \For{each $q \in \mathcal{Q}$}
          \State Sample $\{\tau^{(i)}\}_{i=1}^{N} \sim \pi_\theta(\cdot \mid q)$
          \ForAll{$\tau^{(i)}$}
              \State $r_t \gets \delta^f_t + \delta^m_t + \delta^u_t$ \quad $\forall\, t < T_i$
              \State $r_{T_i} \gets r^{\mathrm{F1}} + r^{\mathrm{DC}}$
              \State $G_{T_i+1} \gets 0$
              \For{$t = T_i$ \textbf{down to} $1$}
                  \State $G_t \gets r_t + \gamma\, G_{t+1}$
              \EndFor
          \EndFor     
          \State $A(\tau^{(i)}) \gets \dfrac{R(\tau^{(i)}) - \mathrm{mean}_j\,
  R(\tau^{(j)})}{\sigma_{\mathcal{G}_\tau} + \epsilon}$
          \ForAll{turn $t$ in each $\tau^{(i)}$}
              \For{$K = 1$ \textbf{to} $H$}
                  \State $\mathcal{G}^{(K)}_t \gets$ turns sharing $\phi_K(t)$
                  \State $\hat{A}^{(K)}_t \gets \dfrac{G_t - \mathrm{mean}_{j \in
  \mathcal{G}^{(K)}_t} G_{t_j}}{\sigma_{\mathcal{G}^{(K)}_t} + \epsilon}$
              \EndFor
              \State $A_t \gets \omega_0 A(\tau^{(i)}) + \sum_{K=1}^{H} \omega_K \hat{A}^{(K)}_t$
              \State $w_t \gets f(A_t)$ \Comment{Eq.~\eqref{eq:weight}}
          \EndFor
          \State $\mathcal{D} \gets \mathcal{D} \cup \{(\tau^{(i)}, \{w_t\})\}_{i=1}^{N}$
      \EndFor
      \State $\theta \gets \arg\min_\theta\, \mathcal{L}(\theta;\, \mathcal{D})$
  \Until{maximum iterations reached}
  \State \Return $\theta$
  \end{algorithmic}
  \end{algorithm}

\begin{figure}[t]
    \centering
    \includegraphics[width=\linewidth]{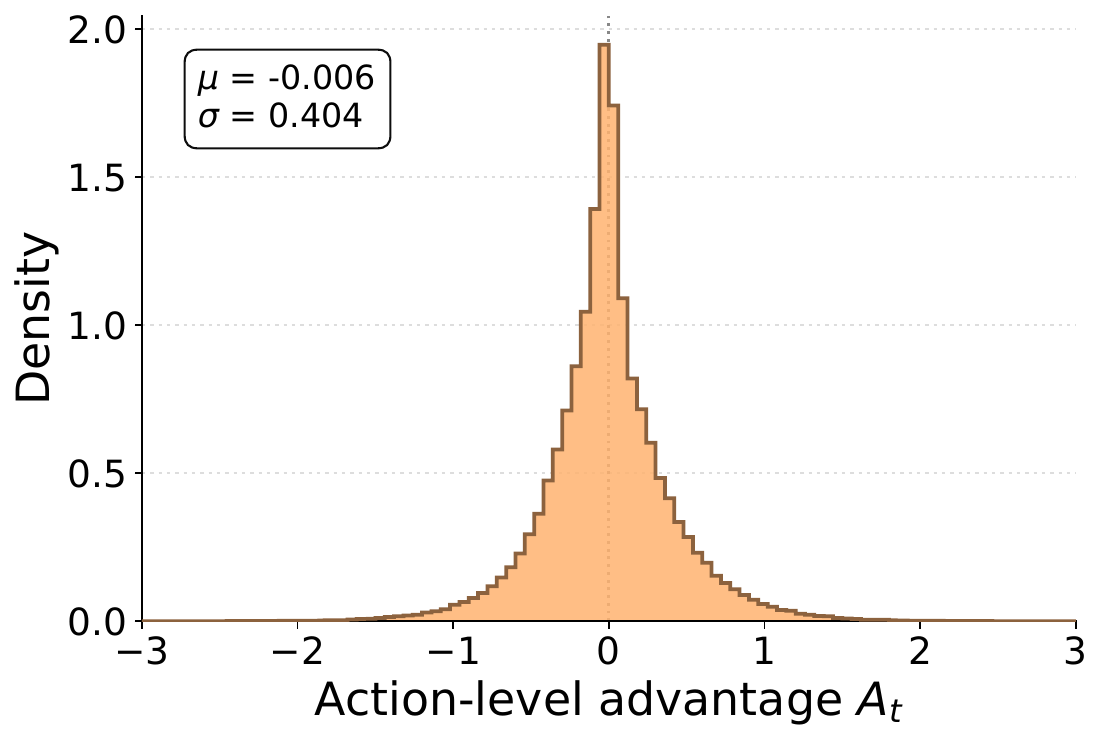}
    \caption{Distribution of per-turn action-level advantage $A_t$
    computed with history depth $H{=}2$ on the training set.
    The distribution is approximately zero-centered
    ($\mu = -0.006$, $\sigma = 0.404$), indicating that
    action-history grouping yields a stable and unbiased baseline
    for action-level credit assignment.}
    \label{fig:advantage_dist}
\end{figure}

\section{Algorithm} \label{app:algorithms}
Algorithm~\ref{alg:training} summarizes the training procedure of our
method. We further report the distribution of action-level
advantage $A_t$ in Figure~\ref{fig:advantage_dist}, showing that the
estimated advantages are well-centered and stable across turns.

\section{Benchmark Details}
\label{app:benchmark}

SWE-Bench~\cite{jimenez2024swebench} is a benchmark for evaluating LLMs
on real-world software engineering tasks, where each instance consists of
a GitHub issue and its corresponding pull request from popular open-source
Python repositories. For our localization task, gold targets at file, module, and function granularities are extracted by parsing files, classes, and functions in the gold patch.

\paragraph{SWE-Bench Verified~{\normalfont\cite{jimenez2024swebench}}.}
A human-validated subset of 500 instances from SWE-Bench, created in
collaboration with OpenAI to improve reliability. It serves as
the de-facto standard for repository-level issue resolution.

\paragraph{SWE-Bench Pro~{\normalfont\cite{deng2025swe}}.}
A harder benchmark sourced from 41 GPL-licensed and commercial
repositories to mitigate data contamination, featuring long-horizon tasks
with an average of 107.4 lines of edits across 4.1 files per instance.

\section{Baseline Methods}
\label{app:baselines}

We describe the baseline methods used for comparison. The tool
configurations of each method are summarized in Table~\ref{tab:tool-comparison}.

\paragraph{Agentless~{\normalfont\cite{xia2024agentless}}.}
A fixed-pipeline method that sequentially executes three stages via LLM
prompting, namely suspicious file candidate extraction, class and function
identification, and line-level localization, relying solely on
in-context reasoning without an agent loop.

\paragraph{CoSIL~{\normalfont\cite{jiang2025cosil}}.}
Incrementally constructs a module call graph through function call analysis
during exploration, expanding from an entry point toward suspicious regions.
Employs context pruning for token budget management.

\paragraph{OrcaLoca~{\normalfont\cite{yu2025orcaloca}}.}
Navigates the repository via priority-based action scheduling, progressively
decomposing file- and class-level search into function-level candidates with
relevance scores. Uses distance-aware context pruning on the code graph to
efficiently narrow down bug locations.

\paragraph{LocAgent~{\normalfont\cite{chen2025locagent}}.}
Indexes the repository as a heterogeneous graph prior to inference,
enabling the agent to search, traverse, and inspect code entities
on the graph. Our tool infrastructure directly follows this method.

\paragraph{RepoSearcher~{\normalfont\cite{ma2025tool}}.}
Adopts a two-stage workflow that progressively narrows from file-level to
function-level localization, and is the first to introduce a distillation-based
training framework (\emph{ToolTrain}) for code localization agents using
Claude 3.7 Sonnet as the teacher model.

\paragraph{RepoNavigator~{\normalfont\cite{zhang2025one}}.}
Equips the agent with a single \texttt{Jump} tool to trace Python symbol
definitions, and trains directly from a pretrained model via GRPO without
relying on any closed-source teacher model.

\paragraph{CodeScout~{\normalfont\cite{sutawika2026codescout}}.}
Adopts a minimal toolset consisting of a standard Unix terminal and
\texttt{LocalizationFinish}, and trains the policy with reinforcement
learning using GSPO.

\section{Evaluation Protocol}
\label{app:eval_protocol}
We formally define the evaluation protocol described in Section~\ref{sec:exp-setup}. Let $\hat{Y}_g$ denote the set of predicted locations and $Y_g$ the set of gold locations at granularity $g \in \{F, M, U\}$. For each instance we compute the scores from the overlap of the two sets and then average them over all instances.
\begin{align}
\mathrm{Recall} &= \frac{|\hat{Y}_g \cap Y_g|}{|Y_g|}, \quad
\mathrm{Precision} = \frac{|\hat{Y}_g \cap Y_g|}{|\hat{Y}_g|}, \nonumber \\
\mathrm{F1} &= \frac{2\,|\hat{Y}_g \cap Y_g|}{|\hat{Y}_g| + |Y_g|}
\end{align}
Importantly, the denominator of precision is not the set of all functions in the repository but the candidate set $|\hat{Y}_g|$ that the model itself submits, and the remaining entities in the repository never enter the score computation. There is therefore no negative class, and the notion of a positive-to-negative ratio is not defined under this protocol.

\section{Implementation Details}
\label{app:impl}
 
All training stages are conducted on 4 NVIDIA RTX Pro 6000 GPUs
(96\,GB VRAM each). All experiments are implemented using the PyTorch
framework with Python v3.12, using HuggingFace Transformers with
DeepSpeed ZeRO-3~\cite{rajbhandari2020zero}. Rollout collection is
served with vLLM~\cite{kwon2023efficient}. Hyperparameters specific to our
method, including history depth $H$, are validated through ablation
studies in Section~\ref{sec:ablation}. Basic training hyperparameters
such as learning rate, batch size, and LoRA configuration are set to
values commonly used in the literature~\cite{hu2021lora,shao2024deepseekmath}.
Detailed hyperparameters are provided in Table~\ref{tab:hparams}.

\paragraph{Teacher SFT Warm-up.}
Using a teacher model\footnote{\texttt{Qwen3-30B-A3B-Instruct-2507}},
we filter 1,410 trajectories that match at least one gold location at
each granularity. Training uses a learning rate of $2{\times}10^{-4}$
and sequence length of 36K. Loss is computed only on assistant turns.

\paragraph{Rollout Collection.}
Trajectories are sampled with temperature 1.0, top-$p$ 0.8, and
repetition penalty 1.05. The maximum number of tool calls per
trajectory is capped at 20, and the maximum context length is 40K. All evaluation results are obtained from a single run with greedy decoding, consistent with the evaluation protocol of prior work.

\paragraph{Advantage-Weighted Reinforcement Learning.}
Token weights are derived from $A_t$ as described in Section~\ref{sec:training}. Eq.~\eqref{eq:loss} is not a clipped policy-gradient objective as in PPO or GRPO but a maximum-likelihood objective weighted by advantages derived from the environment reward, which falls under advantage-weighted training~\cite{peng2019advantage}. Our procedure iterates rollout collection and training updates, re-collecting rollouts from the current policy at each iteration~\cite{gulcehre2023reinforced}, and advantage-weighted training is known to be robust to the resulting mild off-policy drift without importance-ratio correction~\cite{peng2019advantage, nair2020awac}.

The token-weight temperature $\beta$ and advantage-clip bound $\kappa$ are tuned by grid search over $\beta \in \{0.5, 1.0, 1.5, 2.0\}$ and $\kappa \in \{3.0, 5.0, 10.0\}$, with $\beta{=}1.5$ and $\kappa{=}5.0$ selected by validation F1. The policy is updated for up to 3 iterations, each consisting of rollout collection followed by a training update. Each training update runs for 2 epochs
with batch size 16 for 2,046 steps, a learning rate of
$5{\times}10^{-5}$, sequence length 40K, and AdamW optimizer (weight
decay 0.01, 10 warmup steps, linear schedule). LoRA configuration
follows that of Teacher SFT Warm-up.

\paragraph{Training Cost.}

We report the wall-clock time on 4 NVIDIA RTX Pro 6000 GPUs for teacher SFT and for rollout collection with advantage-weighted training per iteration in Table~\ref{tab:training_cost}.

\begin{table}[t]
\centering
\begin{adjustbox}{max width=\linewidth}
\begin{tabular}{llc}
\toprule
\textbf{Model} & \textbf{Stage} & \textbf{Wall-clock} \\ \midrule
\multirow{2}{*}{Qwen3-4B-Instruct} & Teacher SFT & 2.1h \\
 & Rollout + Advantage-weighted training & 2d 6h \\ \midrule
\multirow{2}{*}{Qwen3-8B} & Teacher SFT & 2.6h \\
 & Rollout + Advantage-weighted training & 3d \\ \bottomrule
\end{tabular}
\end{adjustbox}
\caption{Training cost of \textsc{$\text{A}^2$Agent} measured in wall-clock time on 4 NVIDIA RTX Pro 6000 GPUs.}
\label{tab:training_cost}
\end{table}

\paragraph{Token Consumption.}

Since the number of tool calls does not fully reflect the prompt size and generation per call, we report the average input and output tokens per instance over the 500 SWE-Bench Verified instances in Table~\ref{tab:token_consumption}. Input tokens denote the amount the model reads as the prompt, and output tokens denote the amount the model generates. The input tokens of \textsc{$\text{A}^2$Agent} are comparable to those of the multi-turn exploration baselines, whereas the output tokens are among the smallest. This is because our method performs only the necessary actions and finishes exploration in relatively few turns, concentrating on a specific area in the later turns, which is consistent with the shift toward targeted search reported in Section~\ref{sec:efficiency}.

\begin{table}[t]
\centering
\begin{adjustbox}{max width=\linewidth}
\begin{tabular}{lcccc}
\toprule
\textbf{Method} & \textbf{4B Input} & \textbf{4B Output} & \textbf{8B Input} & \textbf{8B Output} \\ \midrule
LocAgent & 77.5K & 3.6K & 178.6K & 1.2K \\
CoSIL & 62.3K & 3.2K & 136.1K & 1.4K \\
RepoSearcher & 108.7K & 1.8K & 107.9K & 1.3K \\
Agentless & 8.2K & 0.3K & 8.2K & 0.2K \\
\textsc{$\text{A}^2$Agent} & 76.8K & 0.9K & 129.1K & 0.9K \\ \bottomrule
\end{tabular}
\end{adjustbox}
\caption{Average input and output tokens per instance on SWE-Bench Verified under the 4B and 8B backbones.}
\label{tab:token_consumption}
\end{table}

\paragraph{Repository Index Construction.}

Our method builds a dependency graph and a BM25 index once per repository snapshot, taking 4.7GB with 1h56m and 2.4GB with 29m respectively over all 500 SWE-Bench Verified instances, and once built they are reused through inference. Such preprocessing is not unique to our method, as graph-based agents also parse repository structure in advance. Moreover, since our static index is read-only, all rollouts share the same index, requiring no execution environment or state restoration per rollout.

\section{Case Study}
\label{app:case_study}

In this section, we provide case studies comparing \textsc{$\text{A}^2$Agent} and the
model trained with only SFT on the same issue,
\texttt{sphinx-doc/sphinx\#10449}.
Figures~\ref{fig:case_ours_1}--\ref{fig:case_ours_3} illustrate the
trajectory of \textsc{$\text{A}^2$Agent}, which successfully localizes the gold
location through appropriate tool calls. In contrast, as shown in
Figure~\ref{fig:case_sft}, the model trained with only SFT fails to
reach the correct location, as it is limited to imitating teacher
trajectories and cannot adapt its search strategy to unseen codebases
effectively. The system prompt used for SFT, rollout generation, and
inference is unified and provided in
Figures~\ref{fig:system-prompt-1}--\ref{fig:system-prompt-3}.

\begin{figure*}[t]\small\raggedright
\issueline\\[2pt]
\textbf{Issue: sphinx-doc/sphinx\#10449}\\
\texttt{autodoc\_typehints="description"} makes \texttt{autoclass} render a spurious
\emph{Return type} for a class (e.g.\ \texttt{class Square} whose \texttt{\_\_init\_\_}
returns \texttt{None}).\\
\textbf{Gold:}\;\texttt{sphinx/ext/autodoc/typehints.py}\;/\;
module \texttt{typehints}\;/\;function \texttt{merge\_typehints}\\[-2pt]
\issueline

\vspace{4pt}
\noindent\textbf{Action 1:} \texttt{SearchSummary(query="sphinx autoclass return type", top\_k=5)}\\
\textbf{Observation 1:}
\begin{lstlisting}[style=obs]
Found 5 files matching 'sphinx autoclass return type':
  doc/.../examples/helloworld.py  - custom "Hello World!" directive (HelloWorld)
  sphinx/highlighting.py          - Pygments syntax highlighting (PygmentsBridge)
  sphinx/domains/__init__.py      - domains: ObjType, Domain
  sphinx/search/tr.py             - Turkish search stemming
  sphinx/util/images.py           - image size / mimetype utils
(none is the autodoc typehints file)
\end{lstlisting}
\turnline

\noindent\textbf{Action 2:} \texttt{ViewSummary(file\_path="sphinx/domains/\_\_init\_\_.py")}\\
\textbf{Observation 2:}
\begin{lstlisting}[style=obs]
Summary for sphinx/domains/__init__.py:
Defines the structure/behavior of domains (group directives & roles).
Key classes: ObjType (object types in a domain), Domain (manages
object types, directives, roles, indices).
\end{lstlisting}
\turnline

\noindent\textbf{Action 3:}
\texttt{TraverseGraph(start=["sphinx/domains/\_\_init\_\_.py"],}
\texttt{direction="downstream", depth=2, dep=["imports","invokes"])}\\
\textbf{Observation 3:}
\begin{lstlisting}[style=obs]
sphinx/domains/__init__.py
|-- imports -- sphinx/addnodes.py:pending_xref
|-- imports -- sphinx/roles.py:XRefRole
|-- imports -- sphinx/roles.py:XRefRole.result_nodes
|-- imports -- sphinx/util/typing.py
|              |-- imports -- sphinx/ext/autodoc/mock.py:ismock
|              |-- imports -- sphinx/ext/autodoc/mock.py:_MockObject
|-- imports -- sphinx/builders/__init__.py:Builder
|-- imports -- sphinx/builders/__init__.py:Builder.get_target_uri
|-- imports -- sphinx/environment/__init__.py:BuildEnvironment
|-- imports -- sphinx/locale/__init__.py
|-- imports -- sphinx/util/docutils.py:SphinxDirective
|-- imports -- sphinx/util/typing.py:RoleFunction
|-- imports -- sphinx/util/typing.py:TitleGetter
|-- imports -- sphinx/directives/__init__.py:ObjectDescription
|-- imports -- sphinx/directives/__init__.py:ObjectDescription.needs_arglist
                ... (16 of 36 lines omitted)
\end{lstlisting}
\issueline

\caption{Example trajectory of \textsc{$\text{A}^2$Agent}, which successfully localizes the gold location. Continued in Figure~\ref{fig:case_ours_3}.}
\label{fig:case_ours_1}
\end{figure*}

\begin{figure*}[t]\small\raggedright
\issueline

\vspace{4pt}
\noindent\textbf{Action 4:} \texttt{SearchCodeSnippet(search\_terms=["autoclass", "autodoc\_typehints"])}\\
\textbf{Observation 4:}
\begin{lstlisting}[style=obs]
[bm25] sphinx/ext/autosectionlabel.py:setup
  51 def setup(app: Sphinx) -> Dict[str, Any]:
  52   app.add_config_value('autosectionlabel_prefix_document', False, 'env')
[bm25] sphinx/ext/autodoc/mock.py:_make_subclass
                ... (~100 of 108 lines omitted)
\end{lstlisting}
\turnline

\noindent\textbf{Action 5:}
\texttt{SearchEntity(entity\_names=["sphinx/ext/autodoc/typehints.py:merge\_typehints"])}\\
\textbf{Observation 5:}
\begin{Verbatim}[fontsize=\small,breaklines=true,
  highlightlines={5,19,21},
  highlightcolor=yellow!30]
Found function sphinx/ext/autodoc/typehints.py:merge_typehints  (exact match)
37 def merge_typehints(app, domain, objtype, contentnode) -> None:
38   if domain != 'py':
39     return
40   if app.config.autodoc_typehints not in ('both', 'description'):   # <-- gold
41     return
43   try:
44     signature = cast(addnodes.desc_signature, contentnode.parent[0])
45     if signature['module']:
46       fullname = '.'.join([signature['module'], signature['fullname']])
47     else:
48       fullname = signature['fullname']
49   except KeyError:
50     return
53   annotations = app.env.temp_data.get('annotations', {})
54   if annotations.get(fullname, {}):
55     field_lists = [n for n in contentnode if isinstance(n, nodes.field_list)]
60     for field_list in field_lists:
61       if app.config.autodoc_typehints_description_target == "all":       # <-- gold
62         modify_field_list(field_list, annotations[fullname])
63       elif ... == "documented_params":                                   # <-- gold
64         augment_descriptions_with_types(field_list, ..., force_rtype=True)
\end{Verbatim}
\issueline

\noindent\textbf{Action 6:}
\texttt{SearchCodeSnippet(search\_terms=["autodoc\_typehints\_description\_target"])}\\
\textbf{Observation 6:}
\begin{lstlisting}[style=obs]
[bm25] -> sphinx/ext/autodoc/typehints.py:merge_typehints  (confirms target)
40 if app.config.autodoc_typehints not in ('both','description'): return
                ... (~90 of 98 lines omitted)
\end{lstlisting}
\turnline

\noindent\textbf{Finish}\\
\textbf{File:}\;\textcolor{green!45!black}{\texttt{sphinx/ext/autodoc/typehints.py}}\;\ding{51}\\
\textbf{Module:}\;\textcolor{green!45!black}{\texttt{typehints}}\;\ding{51}\\
\textbf{Function:}\;\textcolor{green!45!black}{\texttt{merge\_typehints}}\;\ding{51}
\issueline

\caption{Example trajectory of \textsc{$\text{A}^2$Agent}, which successfully localizes the gold location.}
\label{fig:case_ours_3}
\end{figure*}

\begin{figure*}[t]\small\raggedright
\issueline\\[2pt]
\textbf{Issue: sphinx-doc/sphinx\#10449}\\
\texttt{autodoc\_typehints="description"} makes \texttt{autoclass} render a spurious
\emph{Return type} for a class (e.g.\ \texttt{class Square} whose \texttt{\_\_init\_\_}
returns \texttt{None}).\\
\textbf{Gold:}\;\texttt{sphinx/ext/autodoc/typehints.py}\;/\;
module \texttt{typehints}\;/\;function \texttt{merge\_typehints}\\[-2pt]
\issueline

\vspace{4pt}
\noindent\textbf{Action 1:}
\texttt{TraverseGraph(start=['sphinx'], depth=2, dep=['contains'])}\\
\textbf{Observation 1:}
\begin{lstlisting}[style=obs]
sphinx
|-- contains -- sphinx/__init__.py
|-- contains -- sphinx/addnodes.py
|               |-- sphinx/addnodes.py:desc_signature
|               |-- sphinx/addnodes.py:desc_content
                ... (~200 of 206 lines omitted; ext/autodoc/typehints.py not surfaced)
\end{lstlisting}
\turnline

\noindent\textbf{Action 2:} \texttt{SearchSummary(query="autodoc typehints")}\\
\textbf{Observation 2:}
\begin{lstlisting}[style=obs]
No files found matching 'autodoc typehints'.
\end{lstlisting}
\turnline

\noindent\textbf{Action 3:} \texttt{SearchSummary(query="autodoc")}\\
\textbf{Observation 3:}
\begin{lstlisting}[style=obs]
Found 7 files matching 'autodoc':
  doc/conf.py
  doc/.../examples/autodoc_intenum.py
  sphinx/ext/autodoc/__init__.py
    Classes: Documenter, ModuleDocumenter, ClassDocumenter, ExceptionDocumenter, ...
    Functions: identity, members_option, ..., setup
  sphinx/ext/autodoc/directive.py    (AutodocDirective)
  sphinx/ext/autodoc/importer.py     (import_object, get_class_members)
  sphinx/ext/autodoc/mock.py         (_make_subclass, ismock)
  sphinx/ext/autodoc/preserve_defaults.py
(typehints.py is NOT among the results)
\end{lstlisting}
\turnline

\noindent\textbf{Action 4:}
\texttt{ViewSummary(file\_path="sphinx/ext/autodoc/\_\_init\_\_.py")}\\
\textbf{Observation 4:}
\begin{lstlisting}[style=obs]
Summary for sphinx/ext/autodoc/__init__.py:
Classes: Documenter, ModuleDocumenter, ClassDocumenter, ExceptionDocumenter, ...
Functions: identity, members_option, ..., autodoc_attrgetter, setup
Imports: ..., sphinx/ext/autodoc/importer.py, .../mock.py | Edits: 5 commits
(merge_typehints / typehints.py never appears)
\end{lstlisting}
\turnline

\noindent\textbf{Action 5:}
\texttt{SearchEntity(entity\_names=["sphinx/ext/autodoc/\_\_init\_\_.py"])}\\
\textbf{Observation 5:}
\begin{lstlisting}[style=obs]
Found file sphinx/ext/autodoc/__init__.py  (exact match)
   1 """Extension to create automatic documentation from code docstrings.
   8 import re
  12 from typing import (TYPE_CHECKING, Any, Callable, Dict, Iterator, List, ...)
                ... (~200 of 205 lines omitted)
\end{lstlisting}
\turnline

\noindent\textbf{Finish}\\
\textbf{File:}\;\textcolor{red!70!black}{\texttt{sphinx/ext/autodoc/\_\_init\_\_.py}}\;\ding{55}\\
\textbf{Module:}\;\textcolor{red!70!black}{\texttt{autodoc}}\;\ding{55}\\
\textbf{Function:}\;\textcolor{red!70!black}{--}\;\ding{55}\quad
(\texttt{typehints.py} not reached)
\issueline

\caption{Example trajectory of the model trained with only SFT, which fails to localize the gold location.}
\label{fig:case_sft}
\end{figure*}

\begin{figure*}[p]
\centering
\begin{promptbox}{System Prompt for Localization Agent (continued on next page)}
Given the following GitHub problem description, your objective is to localize the *exact* code locations that need modification at THREE granularities:
  1. *File* --- the source file containing the bug or new behavior.
  2. *Module* --- the class (or top-level module-scope function) inside that file.
  3. *Function/Entity* --- the specific method/function that must be edited.

You MUST report all three granularities in your final answer. Reporting only file paths is INSUFFICIENT --- module and function predictions are evaluated independently and contribute to your score.

Follow these steps to localize the issue:

## Step 1: Categorize and Extract Key Problem Information
- Classify the problem statement into the following categories:
  Problem description, error trace, code to reproduce the bug, and additional context.
- Identify candidate modules in the '{package_name}' package mentioned in each category.
- Use extracted keywords and line numbers to search for relevant code references for additional context.

## Step 2: Locate Referenced Modules and Functions
- Accurately determine specific files -> modules -> functions.
- Explore the repository to familiarize yourself with its structure.
- Analyze the described execution flow to identify specific classes / methods being referenced.
- Inspect class hierarchies and method signatures to pin down the exact function.
- Pay special attention to distinguishing between modules with similar names using context and described execution flow.

Output format for each location:
- File only:                'file_path'
- Module (class):           'file_path:ClassName'
- Function (method/entity): 'file_path:ClassName.method_name' or 'file_path:top_level_function'

Example: for method `calculate_sum` of class `MathUtils` in `src/helpers/math_helpers.py`:
- file:    src/helpers/math_helpers.py
- module:  src/helpers/math_helpers.py:MathUtils
- entity:  src/helpers/math_helpers.py:MathUtils.calculate_sum
\end{promptbox}
\caption{System prompt provided to the localization agent.}
\label{fig:system-prompt-1}
\end{figure*}

\begin{figure*}[p]
\centering
\begin{promptbox}{System Prompt for Localization Agent (continued on next page)}
## Step 3: Analyze and Reproduce the Problem
- Clarify the purpose of the issue.
- If expanding capabilities: identify where and how to incorporate new behavior, fields, or modules.
- If addressing unexpected behavior: focus on localizing modules containing potential bugs.

- Reconstruct the execution flow:
  - Identify main entry points triggering the issue.
  - Trace function calls, class interactions, and sequences of events.
  - Identify potential breakpoints causing the issue.
  - Keep the reconstructed flow focused on the problem and avoid irrelevant details.

## Step 4: Locate Areas for Modification (file -> module -> function)
- For every candidate file, ALSO determine WHICH class and WHICH method inside it require changes.
  Do not stop at the file level.
- Use `SearchEntity` to read the actual class/method body and verify the function-level location.
- Consider upstream and downstream dependencies that may affect or be affected by the issue.
- If applicable, identify where to introduce new fields, functions, or variables.
- Think thoroughly: list multiple potential solutions and consider edge cases that could impact the resolution.

## Available Tools and When to Use Them
You have 6 tools. Choose the most appropriate tool for each step. Do NOT repeatedly call the same tool.

### Search Tools
- `TraverseGraph`: first step --- understand repo layout and module hierarchy.
- `SearchSummary`: semantic file discovery.
- `SearchCodeSnippet`: find code by keyword.
- `SearchCommit`: find related commits and changed files.

### View Tools
- `SearchEntity`: read actual code once you know the file/class/function path.
- `ViewSummary`: view the LLM-generated summary of a specific file.
\end{promptbox}
\caption{System prompt provided to the localization agent.}
\label{fig:system-prompt-2}
\end{figure*}

\begin{figure*}[p]
\centering
\begin{promptbox}{System Prompt for Localization Agent}
### Finish
- `Finish`: submit final answer after you have localized all relevant locations.

### Recommended workflow
1. `TraverseGraph` -> get repository overview
2. `SearchSummary` -> find relevant files by concept
3. `ViewSummary` -> understand key files' purposes
4. `SearchCommit` -> find related change history and co-changed files
5. `SearchCodeSnippet` -> locate specific code regions by keyword
6. `SearchEntity` -> read actual code of key functions/classes
7. `Finish` -> output your final answer

## Output Format for Final Results
Your final output should list the locations requiring modification, wrapped with triple backticks.
For EACH location you MUST report all available granularities:
- the file path
- the class name (module-level), if the change is inside a class
- the function name (entity-level): either `ClassName.method` or top-level `function_name`
- line numbers when known

Order locations by importance.
Aim for about 5 files.
Within each file, list every relevant module/function, not just one.

Module-level and function-level predictions are scored independently from file-level.
Reporting only the file path will lose about 2/3 of the available credit.
Always include `class:` and `function:` when the answer lies inside a class or specific function.

### Examples
full_path1/file1.py
line: 10
class: MyClass1
function: MyClass1.my_function1

full_path2/file2.py
line: 76
class: MyClass2
function: MyClass2.my_function2

full_path3/file3.py
line: 24
line: 156
function: my_function3

Return just the location(s).
\end{promptbox}
\caption{System prompt provided to the localization agent.}
\label{fig:system-prompt-3}
\end{figure*}

\end{document}